\documentclass[letterpaper]{article} 
\usepackage{aaai2026}  
\usepackage{times}  
\usepackage{helvet}  
\usepackage{courier}  
\usepackage[hyphens]{url}  
\usepackage{graphicx} 
\usepackage{natbib}  
\usepackage{caption} 
\usepackage{algorithm}
\usepackage{algorithmic}
\usepackage{amsmath}
\usepackage{amssymb}
\usepackage{amsthm}
\usepackage{booktabs}
\usepackage{multirow}
\usepackage{caption}
\usepackage{adjustbox}
\usepackage{xcolor}
\usepackage{graphicx}   
\usepackage{subcaption} 
\usepackage{times}
\usepackage{helvet}
\usepackage{courier}
\usepackage{xcolor}
\newtheorem{theorem}{Theorem}

\newcommand{\x}{{\boldsymbol{x}}}
\newcommand{\vv}{{\boldsymbol{v}}}

\newcommand{\s}{{\boldsymbol{s}}}
\newcommand{\w}{{\boldsymbol{w}}}

\newcommand{\Gauss}{\mathcal{N}}

\newcommand{\eye}{\mathbf{I}}
\newcommand{\dd}{{\rm d}}
\usepackage{newfloat}
\usepackage{listings}

\DeclareCaptionStyle{ruled}{labelfont=normalfont,labelsep=colon,strut=off} 
\floatstyle{ruled}
\newfloat{listing}{tb}{lst}{}
\floatname{listing}{Listing}
\title{A-FloPS: Accelerating Diffusion Models via Adaptive Flow Path Sampler}
\author{
    Cheng Jin,
    Zhenyu Xiao,
    Yuantao Gu\thanks{Corresponding authors.}
}
\affiliations{
    Department of Electronic Engineering\\
Tsinghua University\\
Beijing 100084\\
P. R. China\\
    jinc21@mails.tsinghua.edu.cn, xzy23@mails.tsinghua.edu.cn, gyt@tsinghua.edu.cn
}

\begin{document}

\maketitle

\begin{abstract}
Diffusion models deliver state-of-the-art generative performance across diverse modalities but remain computationally expensive due to their inherently iterative sampling process. Existing training-free acceleration methods typically improve numerical solvers for the reverse-time ODE, yet their effectiveness is fundamentally constrained by the inefficiency of the underlying sampling trajectories.
We propose A-FloPS (Adaptive Flow Path Sampler), a principled, training-free framework that reparameterizes the sampling trajectory of any pre-trained diffusion model into a flow-matching form and augments it with an adaptive velocity decomposition. The reparameterization analytically maps diffusion scores to flow-compatible velocities, yielding integration-friendly trajectories without retraining. The adaptive mechanism further factorizes the velocity field into a linear drift term and a residual component whose temporal variation is actively suppressed, restoring the accuracy benefits of high-order integration even in extremely low-NFE regimes.
Extensive experiments on conditional image generation and text-to-image synthesis show that A-FloPS consistently outperforms state-of-the-art training-free samplers in both sample quality and efficiency. Notably, with as few as $5$ function evaluations, A-FloPS achieves substantially lower FID and generates sharper, more coherent images. The adaptive mechanism also improves native flow-based generative models, underscoring its generality. These results position A-FloPS as a versatile and effective solution for high-quality, low-latency generative modeling.
\end{abstract}


\section{Introduction}

Diffusion models have achieved remarkable generative performance across diverse modalities, producing high‑fidelity samples that often rival real‑world data~\cite{rombach2022high,esser2024scaling,ramesh2022hierarchical}.  
However, their practical deployment is hindered by substantial sampling latency due to the inherently iterative nature of the generation process.  
Producing a single high‑quality sample typically requires hundreds of network evaluations, making inference computationally expensive and limiting applicability in real‑time or resource‑constrained scenarios~\cite{ho2020denoising,song2021scorebased}.

To address this bottleneck, a variety of acceleration strategies have been developed, which can be broadly categorized into two groups.  
The first group comprises \emph{training‑required} approaches, such as knowledge distillation~\cite{salimans2022progressive,schusterbauer2025diff2flow}—which compress a teacher model’s multi‑step generation process into a student model capable of producing comparable results in far fewer steps—and consistency models~\cite{song2023consistency,sauer2023latents}, which train networks to satisfy a self‑consistency constraint across noise levels, enabling high‑quality generation with an arbitrary (and often very small) number of inference steps.  
While highly effective, these methods require costly retraining and their performance is closely tied to the specific base model and training configuration.

The second group consists of \emph{training‑free} methods, which accelerate sampling by improving the numerical integration of the underlying generative ODEs without modifying model parameters.  
Early work such as DDIM~\cite{song2021denoising}—interpretable as a first‑order discretization of the deterministic ODE formulation of diffusion models—showed that this approach outperforms discretizing the stochastic SDE, especially in low‑NFE regimes.  
More recent high‑order solvers like DPM‑Solver~\cite{lu2022dpm,lu2023dpmpp} further improve stability and accuracy, though their performance still degrades when the step count becomes extremely small, leaving room for improvement.

Meanwhile, the \emph{Flow Matching} (FM) framework~\cite{lipman2023flow} has emerged as a compelling paradigm for generative modeling.  
FM learns a continuous‑time velocity field that deterministically transports a simple base distribution to the target data distribution, and often achieves competitive or superior results with fewer discretization steps~\cite{liu2022flow,esser2024scaling}.  
For Gaussian base distributions, FM can be interpreted as a special case of a diffusion model under a particular noise schedule and a deterministic sampling path~\cite{patel2024exploring,gao2025diffusionmeetsflow}.  
However, this correspondence holds only for a specific schedule; a general diffusion model with an arbitrary noise schedule cannot be directly expressed in FM form.

Motivated by this connection, we introduce the \emph{Flow Path Sampler} (FloPS), which reparameterizes the sampling trajectory of an arbitrary pre‑trained diffusion model into a path consistent with FM dynamics.  
This transformation retains the original score network while inheriting the temporal regularity of FM trajectories, enabling more stable and efficient ODE‑based integration—entirely without retraining.

Building on this foundation, we further propose the \emph{Adaptive Flow Path Sampler} (A‑FloPS), which incorporates an adaptive decomposition of the velocity field to minimize the temporal variability of the residual component.  
This adaptivity allows high‑order numerical solvers to operate more effectively, resulting in substantial improvements in both stability and sample quality, particularly in the few‑step regime.

Our work makes the following contributions:

\begin{itemize}
    \item \textbf{Trajectory-level acceleration via diffusion-to-flow reparameterization.}  
    We derive an analytical mapping that transforms the sampling trajectory of \emph{any} pre-trained diffusion model into an integration-friendly flow-matching (FM) form—without retraining or altering network parameters. This enables plug-and-play acceleration for a wide range of generative models.

    \item \textbf{Adaptive velocity decomposition for few-step generation.}  
    We introduce an inference-time adaptive mechanism that factorizes the FM velocity field into a dominant linear drift and a smooth residual, actively suppressing temporal variability. This restores the numerical advantages of high-order integration in the low-NFE regime.

    \item \textbf{Unified framework for diffusion and native FM models.}  
    Beyond reparameterized diffusion models, our adaptive mechanism directly benefits native FM generators, demonstrating its generality across deterministic generative ODEs. 

    \item \textbf{State-of-the-art few-step performance.}  
    Extensive experiments on conditional image generation and text-to-image synthesis show that A-FloPS consistently outperforms strong training-free baselines, delivering sharper details, stronger semantic alignment, and substantially lower FID with as few as 5 steps.
\end{itemize}

\section{Background}

\subsection{Diffusion Model}

Modern diffusion models, originally introduced as discrete-time Markov chains, are now predominantly formulated from the perspective of stochastic differential equations (SDEs)~\cite{song2021scorebased}.  
This continuous-time view facilitates theoretical analysis and allows the use of advanced numerical solvers.  
In this framework, the forward diffusion process gradually perturbs data into noise according to:
\begin{equation}
\label{eq:forwardDM_special}
\dd \x_\tau = -\alpha(\tau)\x_\tau\dd \tau + \sqrt{\beta(\tau)}\dd \w, \quad \x_0 \sim p_0,
\end{equation}
where \( \w \) denotes the Wiener process, \( p_0 \) is the target data distribution, \(\alpha(\tau)\) is the drift coefficient, and \(\beta(\tau)\) is the diffusion coefficient.  

The marginal distribution of \(\x_\tau\) conditioned on \(\x_0\) is:
\begin{equation}
\label{eq:marginal_forward}
\x_\tau \mid \x_0 \sim \mathcal{N}(\bar{\alpha}_\tau \x_0, \sigma^2_\tau \mathbf{I}),
\end{equation}
where
{\small
\[
\sigma^2_\tau = \int_0^\tau \exp\!\left( -2 \int_s^\tau \alpha(r) \, \mathrm{d}r \right) \beta(s) \, \mathrm{d}s\]
}
and
{\small\[
\bar{\alpha}_\tau = \exp\!\left(-\int_0^\tau \alpha(s) \, \mathrm{d}s\right).
\]}
At the terminal time \(\tau = T\), \(\sigma_T^2\) is large and \(\bar{\alpha}_T\) is small, meaning the state approaches an isotropic Gaussian dominated by noise.

The generative process is obtained by reversing the SDE:
\begin{equation}
\label{reverse:sde}
\dd \tilde{\x}_\tau = \left[-\alpha(\tau)\tilde{\x}_\tau - \beta(\tau) \nabla_{\tilde{\x}_\tau} \log p_\tau(\tilde{\x}_\tau)\right] \dd \tau + \sqrt{\beta(\tau)}\dd \tilde{\w},
\end{equation}
where \( \nabla_{\tilde{\x}_\tau} \log p_\tau \) is the \emph{score function} and \( \tilde{\w} \) is the reverse-time Wiener process.  
Alternatively, one can remove the stochastic term to obtain the reverse-time ODE:
\begin{equation}
\label{reverse:ode}
\dd \bar{\x}_\tau = \left[-\alpha(\tau)\bar{\x}_\tau - \tfrac{\beta(\tau)}{2} \nabla_{\bar{\x}_\tau} \log p_\tau(\bar{\x}_\tau)\right] \dd \tau,
\end{equation}
which is preferred in practice due to improved numerical stability and efficiency.

In practice, the exact score function is intractable. Since \(p_T\) is approximately Gaussian, a neural network \( \boldsymbol{s}_\theta(\x_\tau, \tau) \) is trained via denoising score matching to approximate it, and the reverse process is initialized from \(\bar{\x}_T \sim \mathcal{N}(\mathbf{0}, \sigma_T^2\mathbf{I})\).  
Besides direct score prediction, alternative parameterizations—such as noise prediction and velocity prediction—are also widely used~\cite{Karras2022edm}, and can be algebraically converted to the score function for sampling.

\subsection{Accelerated Sampling Methods}

Due to their iterative nature, diffusion model samplers often require hundreds of neural evaluations, leading to high inference latency.  
Acceleration strategies fall into two broad categories: \emph{training‑required} and \emph{training‑free} approaches.

\paragraph{Training‑Required Acceleration.}  
These methods train specialized generative models to reproduce the denoising behavior of a high‑fidelity model on a coarse time grid.  
Knowledge distillation transfers the generative process from a \emph{teacher} to a \emph{student} with fewer steps, as in Progressive Distillation~\cite{salimans2022progressive}, which halves the number of steps iteratively.  
Consistency models~\cite{song2023consistency,sauer2023latents} enforce self‑consistency across noise levels, enabling generation with very few steps.  
Hybrid designs, such as Truncated Diffusion Models~\cite{bao2022truncated}, replace later denoising stages with fast feed‑forward generators.

\paragraph{Training‑Free Acceleration.}  
These methods improve sampling efficiency without retraining by designing better numerical solvers for the reverse ODE.  
DDIM~\cite{song2021denoising} reformulates sampling as a deterministic ODE and applies first‑order integration.  
High‑order solvers like DPM‑Solver~\cite{lu2022dpm} and DPM‑Solver++~\cite{lu2023dpmpp} exploit the semi‑linear structure for higher accuracy in few‑step generation.  
UniPC~\cite{zhao2023unipc} further unifies predictor–corrector schemes for diverse noise schedules, reducing the quality gap to full‑trajectory sampling.

While these approaches improve integration accuracy, they remain constrained by the inherent structure of the sampling trajectory.  
This motivates exploring \emph{trajectory reparameterization}, which can potentially make the path itself more integration‑friendly.

\subsection{Flow Matching Model}

Flow matching~\cite{lipman2023flow} formulates generative modeling as learning a \emph{deterministic} continuous‑time flow that transports a simple reference distribution \( q_0 \) to a complex target distribution \( q_1 \).  
It defines a time‑dependent velocity field \(\vv(\x_t, t)\) evolving samples via:
\begin{equation}
\label{eq:fm-sampling}
\frac{\mathrm{d}\x_t}{\mathrm{d}t} = \vv^*(\x_t, t),
\end{equation}
with \(\x_0 \sim q_0\) and the terminal state \(\x_1\) following \(q_1\).

A common example is Gaussian Optimal Transport, where \(q_0 = \mathcal{N}(\mathbf{0}, \mathbf{I})\) and \(q_1\) is the data distribution.  
The forward interpolation is:
\[
\x_t = (1 - t)\x_0 + t\x_1,
\]
where \(\x_0 \sim q_0\) and \(\x_1 \sim q_1\).  
The optimal velocity field minimizes:
\begin{equation}
\label{eq:fm-training}
\min_{\vv} \, \mathbb{E}_{t,\x_0,\x_1} \big\| \vv(\x_t, t) - (\x_1 - \x_0) \big\|^2,
\end{equation}
whose solution is the conditional expectation:
\begin{equation}
\label{eq:vstar}
\vv^*(\x_t, t) = \mathbb{E}\!\left[\x_1 - \x_0 \,\middle|\, \x_t \right].
\end{equation}
This guarantees a globally consistent transport from \(q_0\) to \(q_1\).

Notably, flow matching can be interpreted as a deterministic sampler variant of diffusion models under a specific noise schedule. 
This connection enables reparameterizing diffusion trajectories into FM‑like paths for potentially more efficient sampling.

\begin{algorithm}[tb]
\caption{Flow Path Sampler(FloPS)}
\label{alg:flops}
\textbf{Input}: the estimation of score function \(\s_\theta(\cdot,\cdot)\), the number of sampling steps \(N\), and the noise scheduler of the diffusion model\(\{\sigma_\tau,\bar \alpha_t|\tau=0,1,2,...,T\}\).\\
\textbf{Output}: Your algorithm's output
\begin{algorithmic}[1] 
\STATE Sample \(\x_0\sim\Gauss(0,\eye)\).
\STATE \(\Delta t\leftarrow 1/N \)
\STATE \( t_n\leftarrow n\Delta t \)
\FOR{\(n=0 \text{ to } N-1\)}
\IF {\(t_n<\frac{1}{1 + \sigma_T / \bar{\alpha}_T}\)}
\STATE \(\vv_{t_n}\leftarrow\frac{{\sigma_T} \left[ \x_t + \sigma_T \s_\theta\left( (\bar{\alpha}_T + \sigma_T)\x_T,T \right) \right]}{\bar{\alpha}_T}\).
\ELSE
\STATE \(\tau\leftarrow\arg\min_{s}\left\lvert t_n-\frac{1}{1 + \sigma_s / \bar{\alpha}_s}\right\rvert \).
\STATE \(\vv_{t_n}\leftarrow\frac{{\sigma_\tau} \left[ \x_t + \sigma_\tau \s_\theta\left( (\bar{\alpha}_\tau + \sigma_T)\x_\tau,\tau \right) \right]}{\bar{\alpha}_\tau(1-t_n)}\).
\ENDIF
\STATE \(\x_{t_{n+1}}\leftarrow \x_{t_n}+\vv_{t_n}\Delta t\)
\ENDFOR
\STATE \textbf{return} \(\x_1\)
\end{algorithmic}
\end{algorithm}

\section{Method}
Our approach introduces a training-free acceleration framework for diffusion model sampling, composed of two key components: \textbf{Flow Path Sampler (FloPS)} and its adaptive variant \textbf{A-FloPS}. 
FloPS reparameterizes the sampling trajectory of any pre-trained diffusion model into a flow-matching (FM) trajectory, enabling more efficient ODE integration without retraining. 
Building on this foundation, A-FloPS further incorporates an \emph{adaptive decomposition} of the velocity field, which suppresses temporal variability in the residual term and thus restores the effectiveness of high-order integration under the FM framework. 
The following subsections detail each component in turn, from the motivation and formulation of FloPS to the design and implementation of our adaptive mechanism.

\subsection{Flow Path Sampler (FloPS)}
Flow Matching (FM) models have been shown to consistently outperform standard diffusion models under the same number of discretization steps, primarily due to their smoother and more regular sampling trajectories induced by the noise schedule~\cite{lipman2023flow}. 
This observation naturally raises the question: \emph{can we reconstruct the sampling trajectory of a diffusion model in an analogous manner, thereby improving both sampling efficiency and generation quality?}

We address this question by deriving an explicit mapping from the score function of a diffusion model to the velocity field of an FM model targeting the same data distribution. 
This mapping is formalized in Theorem~\ref{thm:score-to-velocity}, with a detailed proof provided in the Appendix. 
Intuitively, it enables the conversion of any pre‑trained diffusion model into a valid FM trajectory \emph{without} retraining or architectural modifications.  
From a practical standpoint, this transformation bridges the algorithmic gap between two previously distinct paradigms: score-based stochastic generative modeling and deterministic flow-based modeling. 
It reveals that the efficiency of FM sampling is not an inherent property of its architecture, but rather a consequence of a more integration-friendly trajectory structure, which can in principle be induced from any diffusion model.

\begin{theorem}[Diffusion‑to‑Flow Transformation]
\label{thm:score-to-velocity}
If a diffusion model and a flow‑matching model target the same distribution, i.e., \(q_1\) in Eq.~\eqref{eq:fm-training} equals \(p_0\) in Eq.~\eqref{eq:forwardDM_special}, their velocity field and score function satisfy:
\[
(1 - t)\vv^*(\x_t, t) = \frac{\sigma_\tau}{\bar{\alpha}_\tau} \left[ \x_t + \sigma_\tau \nabla \log p_\tau\!\left( (\bar{\alpha}_\tau + \sigma_\tau)\x_t \right) \right],
\]
where \( t = \frac{1}{1 + \sigma_\tau / \bar{\alpha}_\tau} \) and this transformation is bijective since \(\sigma_\tau / \bar{\alpha}_\tau\) increases strictly with \(\tau\), ensuring a one‑to‑one correspondence between \(\tau\) and \(t\).  
\end{theorem}

Building on this theoretical insight, we propose the Flow Path Sampler (FloPS), a training‑free, plug‑and‑play framework that reuses the pre‑trained score network of a diffusion model to construct its FM trajectory. 
Unlike conventional FM training, FloPS does not require a separately learned velocity field and is applicable to models with \emph{arbitrary} noise schedules or prediction parameterizations.  
This universality is crucial for real-world adoption: FloPS can be seamlessly integrated into existing pipelines, including large text-to-image models and high-resolution conditional generators, without access to their training data or checkpoints beyond the original network weights.

In practice, many diffusion schedulers maintain a non‑zero signal strength at the start of sampling (\(\bar{\alpha}_T \neq 0\)). Consequently, the mapping in Theorem~\ref{thm:score-to-velocity} is exact only after an initial \emph{start time} \(t_{\min}\), given by
\[
t_{\min} = \frac{1}{1 + \sigma_T / \bar{\alpha}_T}.
\]
For earlier times \(t < t_{\min}\), we approximate the velocity using the value at \(t_{\min}\), i.e., \(\vv(\x_t, t_{\min})\). 
The complete FloPS algorithm is summarized in Algorithm~\ref{alg:flops}, where we present the method in its score‑prediction form; equivalent formulations for noise‑ or data‑prediction are straightforward.

\begin{algorithm}[tb]
\caption{Adaptive Flow Path Sampler(A-FloPS)}
\label{alg:aflops}
\textbf{Input}: the estimation of score function \(\s_\theta(\cdot,\cdot)\), the number of sampling steps \(N\), and the noise scheduler of the diffusion model\(\{\sigma_\tau,\bar \alpha_t|\tau=0,1,2,...,T\}\).\\
\textbf{Output}: Your algorithm's output
\begin{algorithmic}[1] 
\STATE Sample \(\x_0\sim\Gauss(0,\eye)\).
\STATE \(\Delta t\leftarrow 1/N \)
\STATE \(t_n = n\Delta t\)
\FOR{\(n=0 \text{ to } N-1\)}
\IF {\(t_n<\frac{1}{1 + \sigma_T / \bar{\alpha}_T}\)}
\STATE \(\vv_{t_n}\leftarrow\frac{{\sigma_T} \left[ \x_t + \sigma_T \s_\theta\left( (\bar{\alpha}_T + \sigma_T)\x_T,T \right) \right]}{\bar{\alpha}_T}\).
\ELSE
\STATE \(\tau\leftarrow\arg\min_{s}\left\lvert t_n-\frac{1}{1 + \sigma_s / \bar{\alpha}_s}\right\rvert \).
\STATE \(\vv_{t_n}\leftarrow\frac{{\sigma_\tau} \left[ \x_t + \sigma_\tau \s_\theta\left( (\bar{\alpha}_\tau + \sigma_T)\x_\tau,\tau \right) \right]}{\bar{\alpha}_\tau(1-t_n)}\).
\ENDIF
\IF {n=0}
\STATE \(\x_{t_{n+1}}\leftarrow \x_{t_n}+\vv_{t_n}\Delta t\)
\ELSE
\STATE Calculate \(\lambda^{(n)}\) as \eqref{eq:soll}
\STATE Calculate \(a,b\) as in ~\eqref{eq:ab}
\STATE Calculate $\x_{t_{n+1}}$ as in~\eqref{eq:main} using the finite-difference form of the derivative in~\eqref{eq:der}
\ENDIF
\ENDFOR
\STATE \textbf{return} \(\x_1\)
\end{algorithmic}
\end{algorithm}

\subsection{Adaptive Mechanism}
High-order samplers such as DPM-Solver++~\cite{lu2022dpm} achieve acceleration primarily through two mechanisms:  
(\emph{i}) employing a high-order expansion of the estimator over the integration interval, thereby improving numerical accuracy; and  
(\emph{ii}) leveraging the approximate \emph{temporal invariance} of the estimator $\mathbb{E}[\x_0 \mid \x_\tau]$, which enables accurate extrapolation from future time steps.  
The second factor is particularly important, as it allows the solver to exploit smooth temporal behavior for better high-order integration.

However, in flow matching, the velocity field $\frac{\dd \x_t}{\dd t}$ exhibits only \emph{weak temporal variation} along the sampling trajectory (e.g., exactly constant for a Dirac delta target), which renders factor~(\emph{ii}) largely ineffective.  
As a result, even a simple Euler integrator can already achieve excellent performance, and the benefits of applying high-order samplers such as DPMSolver diminish considerably.  
This means that naively applying high-order methods to FM-like trajectories can even waste computation, since the additional polynomial fitting captures almost no extra curvature information, which has been found in practice~\cite{tan2025stork}.

To overcome this limitation, we propose an \emph{adaptive decomposition} that factorizes the velocity field into a linear term and a higher-order residual term in an adaptive manner.  
By suppressing the temporal variation of the residual, our method restores the effectiveness of high-order integration within the flow-matching framework.

\paragraph{Adaptive coefficient estimation.}
We rewrite the flow-matching ODE~\eqref{eq:fm-sampling} as:
\begin{equation}
\frac{\dd \x_t}{\dd t} = \lambda_t \x_t + \big[\vv(\x_t, t) - \lambda_t \x_t\big],
\end{equation}
where $\lambda_t$ is a time-dependent coefficient chosen to make the residual $\vv(\x_t, t) - \lambda_t \x_t$ as temporally smooth as possible.  
Unlike the fixed or pre-trained decompositions in the DPM-Solver family, $\lambda_t$ is \emph{adaptive}—it is re-estimated at every integration step to track local variations in the velocity field.  
The intuition is simple: if the remaining term varies slowly, high-order scheme is allowed to exploit temporal smoothness effectively.

For stability and efficiency, we treat $\lambda_t$ as piecewise constant within each integration interval:
\begin{equation}
\label{assum:lambda}
\lambda_t = \lambda^{(n)}, \quad t \in [t_n, t_{n+1}),
\end{equation}
where $0 = t_0 < t_1 < \dots < t_N = 1$.  
Ideally, $\lambda^{(n)}$ minimizes the instantaneous temporal change of the residual:
\begin{equation}
\lambda^{(n)}_{\text{ideal}}
= \arg\min_{\lambda} \left.\left\| \frac{\dd}{\dd t} \boldsymbol{h}(\x_{t}, t; \lambda) \right\|_2^2 \right|_{t = t_n},
\end{equation}
with $\boldsymbol{h}(\x_t, t; \lambda) := \vv(\x_t, t) - \lambda \x_t$.

Direct derivative computation is costly, so we approximate it by a finite difference between consecutive steps:
\begin{equation}
\lambda^{(n)} = \arg\min_{\lambda} \left\| \boldsymbol{h}(\x_{t_n}, t_n; \lambda) - \boldsymbol{h}(\x_{t_{n-1}}, t_{n-1}; \lambda) \right\|_2^2.
\end{equation}
This quadratic problem has the closed-form solution:
\begin{equation}
\label{eq:soll}
\lambda^{(n)} = \frac{\langle \Delta \vv, \Delta \x \rangle}{\lVert \Delta \x \rVert_2^2},
\end{equation}
where $\Delta \vv := \vv(\x_{t_n}, t_n) - \vv(\x_{t_{n-1}}, t_{n-1})$ and $\Delta \x := \x_{t_n} - \x_{t_{n-1}}$.  
This choice minimizes local temporal variation in the residual, improving high-order integrator accuracy.

\begin{figure*}[ht]
    \centering
    \includegraphics[width=0.9\linewidth]{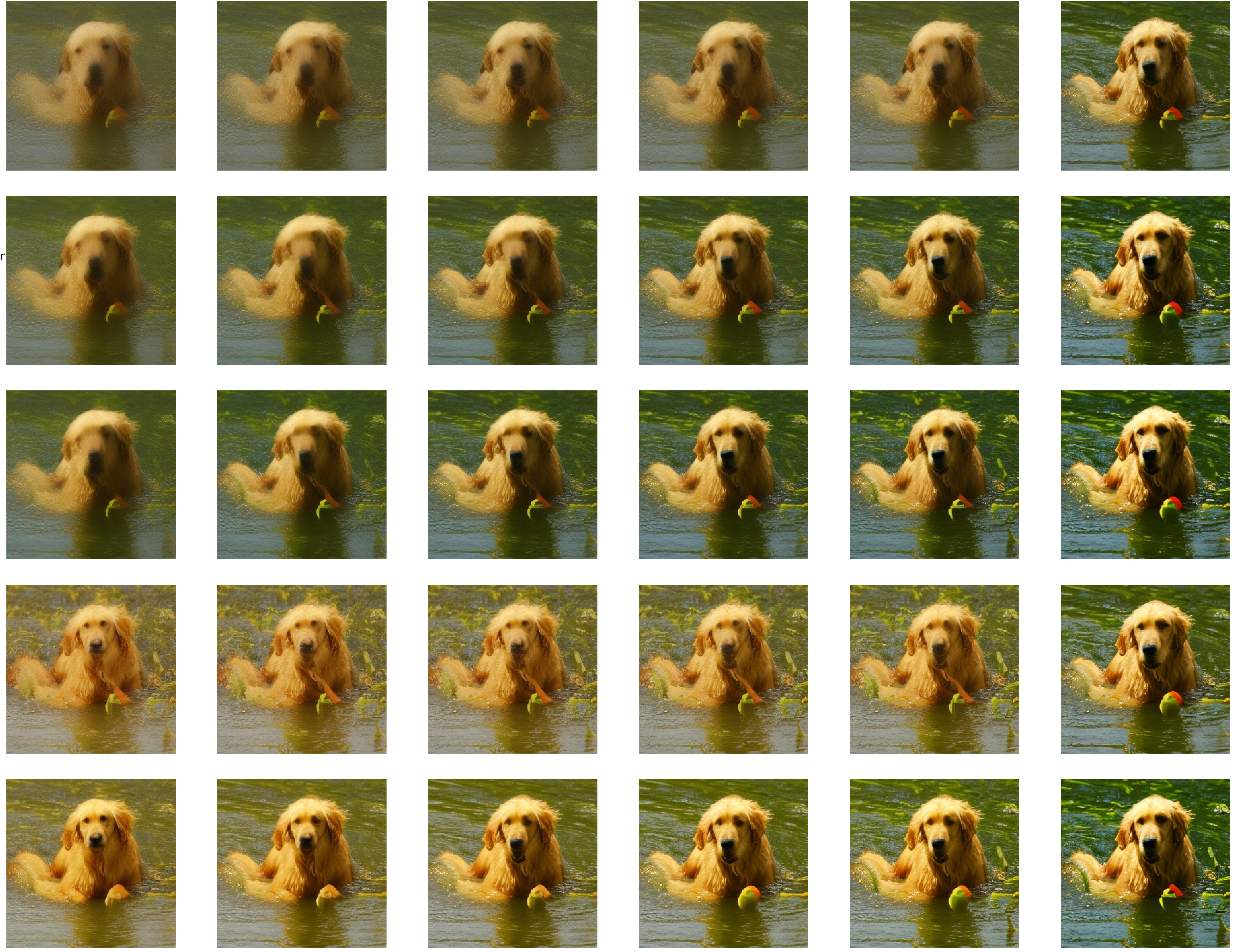}
    \caption{
Qualitative comparison of generated images for the ``golden retriever'' class using different diffusion samplers across varying inference steps.  
Each row corresponds to a specific sampler (from top to bottom: DDIM, DPM-Solver++, UniPC, FloPS, and A-FloPS), while each column shows results at a different number of function evaluations (NFE = 5, 6, 7, 8, 9, 25).  
All methods are initialized with the same random seed to ensure a fair comparison.  
Our proposed A-FloPS generates sharper textures and more coherent object structures, achieving superior visual fidelity with substantially fewer sampling steps.
    }
    \label{fig:qualitative}
\end{figure*}

\paragraph{High-order integration.}
To achieve a better trade-off between discretization error and the number of function evaluations (NFEs), we employ a high-order integration scheme for the flow ODE. 
With~\eqref{assum:lambda}, the flow ODE can be written as:
\begin{align}
\x_{t_{n+1}} &= \mathrm{e}^{\lambda^{(n)} \Delta t} \x_{t_n}\notag\\
&+ \int_{t_n}^{t_{n+1}} \mathrm{e}^{\lambda^{(n)} (t_{n+1} - \tau)} \boldsymbol{h}(\x_\tau, \tau;\lambda^{(n)}) \, \mathrm{d} \tau,
\end{align}
where $\Delta t = t_{n+1} - t_n$.  
Approximating the integral by a second-order Taylor expansion at $t_n$ yields:
\begin{equation}
\label{eq:main}
\begin{split}
\x_{t_{n+1}} &\approx \mathrm{e}^{\lambda^{(n)} \Delta t} \x_{t_n} + a \, \boldsymbol{h}(\x_{t_n}, t_n;\lambda^{(n)}) \\
&\quad + b \left.\frac{\mathrm{d} \boldsymbol{h}(\x_t, t;\lambda^{(n)})}{\mathrm{d}t}\right|_{t = t_n},
\end{split}
\end{equation}
where
\begin{equation}
\label{eq:ab}
a = \frac{1 - \mathrm{e}^{-\lambda^{(n)} \Delta t}}{\lambda^{(n)}}, \quad
b = \frac{1 - (1 + \lambda^{(n)} \Delta t)\mathrm{e}^{-\lambda^{(n)} \Delta t}}{(\lambda^{(n)})^2}.
\end{equation}
In practice, instead of computing the time derivative in~\eqref{eq:main} explicitly, we use its backward finite-difference approximation:
\begin{equation}
\label{eq:der}
\left.\frac{\mathrm{d} \boldsymbol{h}}{\mathrm{d}t}\right|_{t = t_n}
\approx
\frac{\boldsymbol{h}(\x_{t_n}, t_n;\lambda^{(n)}) - \boldsymbol{h}(\x_{t_{n-1}}, t_{n-1};\lambda^{(n)})}{\Delta t}.
\end{equation}
Hence, the update in~\eqref{eq:main} is implemented using the finite-difference estimate~\eqref{eq:der} in place of the time derivative.


This adaptive decomposition is applied entirely at inference time, requiring \emph{no} retraining or architectural changes.  
By continually re-estimating $\lambda_t$, we suppress residual temporal variation and restore the accuracy gains of high-order integration—particularly in the low-step regime where fixed-coefficient methods often underperform.  
We emphasize that this adaptivity is local in time and model-agnostic: it can be applied not only to FloPS but also to any FM-style sampler, and even to certain deterministic ODE solvers outside the generative modeling domain.

\section{Experiments}

We evaluate A-FloPS across diverse generative settings to assess both sampling efficiency and generation quality.  
Our study addresses three main questions:  
(1) How does A-FloPS compare to existing training-free samplers under varying NFEs?  
(2) What is the individual contribution of the proposed flow-based reparameterization?  
(3) How effective is the adaptive mechanism in both diffusion and flow-matching models?

\begin{table*}[ht]
\centering

\begin{adjustbox}{max width=\textwidth}
\begin{tabular}{c | cccc | cccc | cccc | cccc}
\toprule
\multirow{2}{*}{Step} & \multicolumn{4}{c|}{DDIM} & \multicolumn{4}{c|}{UniPC} & \multicolumn{4}{c|}{DPM-Solver} & \multicolumn{4}{c}{A-FloPS (proposed)} \\
\cmidrule(lr){2-5} \cmidrule(lr){6-9} \cmidrule(lr){10-13} \cmidrule(lr){14-17}
 & FID & IS & Prec. & Rec. & FID & IS & Prec. & Rec. & FID & IS & Prec. & Rec. & FID & IS & Prec. & Rec. \\
\midrule
5 & 38.124 & 78.897 & 0.447 & 0.438 & 19.165 & 124.850 & 0.598 & 0.431 & 22.267 & 115.066 & 0.571 & 0.416 & \textbf{6.977} & \textbf{195.176} & \textbf{0.751} & \textbf{0.465} \\
6 & 22.449 & 119.865 & 0.570 & 0.456 & 10.165 & 172.094 & 0.699 & 0.472 & 11.852 & 162.066 & 0.681 & 0.454 & \textbf{4.947} & \textbf{224.613} & \textbf{0.790} & \textbf{0.486} \\
7 & 14.650 & 151.406 & 0.646 & 0.463 & 6.698 & 206.204 & 0.752 & \textbf{0.493} & 7.555 & 196.617 & 0.739 & 0.486 & \textbf{4.196} & \textbf{239.762} & \textbf{0.809} & 0.492 \\
8 & 10.325 & 175.733 & 0.700 & 0.481 & 5.320 & 225.468 & 0.780 & \textbf{0.504} & 5.711 & 216.763 & 0.772 & 0.502 & \textbf{3.753} & \textbf{248.932} & \textbf{0.817} & 0.500 \\
9 & 7.669 & 196.445 & 0.735 & 0.489 & 4.828 & 237.609 & 0.794 & 0.514 & 4.947 & 231.484 & 0.787 & 0.513 & \textbf{3.647} & \textbf{253.016} & \textbf{0.821} & \textbf{0.516} \\
10 & 6.234 & 209.926 & 0.759 & 0.506 & 4.638 & 243.356 & 0.803 & 0.516 & 4.615 & 239.502 & 0.799 & 0.517 & \textbf{3.440} & \textbf{258.486} & \textbf{0.826} & \textbf{0.521} \\
\bottomrule
\end{tabular}
\end{adjustbox}
\caption{Quantitative comparison of different samplers under varying numbers of steps. FID (↓), IS (↑), Precision (↑), and Recall (↑) are reported.}
\label{tab:main_dit}
\end{table*}
\begin{table*}[ht]
\centering

\begin{adjustbox}{max width=\textwidth}
\begin{tabular}{c | cccc | cccc}
\toprule
\multirow{2}{*}{Step} & \multicolumn{4}{c|}{DDIM} & \multicolumn{4}{c}{FloPS (proposed)} \\
\cmidrule(lr){2-5} \cmidrule(lr){6-9}
 & FID & IS & Prec. & Rec. & FID & IS & Prec. & Rec. \\
\midrule
5 & 38.12 & 78.90 & 0.4465 & 0.4381 & 14.9 (–23) & 151.44 (+72.5) & 0.6425 (+0.2) & 0.458 (+0.02) \\
6 & 22.45 & 119.87 & 0.5702 & 0.456 & 9.721 (–13) & 181.12 (+61.3) & 0.7057 (+0.14) & 0.4814 (+0.025) \\
7 & 14.65 & 151.41 & 0.6465 & 0.4631 & 7.005 (–7.6) & 204.47 (+53.1) & 0.7469 (+0.1) & 0.4869 (+0.024) \\
8 & 10.33 & 175.73 & 0.6996 & 0.4813 & 5.587 (–4.7) & 219.57 (+43.8) & 0.7713 (+0.072) & 0.4964 (+0.015) \\
9 & 7.669 & 196.45 & 0.7351 & 0.4889 & 4.623 (–3) & 231.76 (+35.3) & 0.7904 (+0.055) & 0.5042 (+0.015) \\
10 & 6.234 & 209.93 & 0.7591 & 0.5062 & 4.11 (–2.1) & 241.02 (+31.1) & 0.8026 (+0.044) & 0.5052 (–0.001) \\
\bottomrule
\end{tabular}
\end{adjustbox}
\caption{Comparison of DDIM and FloPS across steps. FLoPS values are annotated with color-coded deltas relative to DDIM.}
\label{tab:ablation_flow_only}
\end{table*}

\subsection{Experimental Setup}

We consider two representative generative modeling scenarios: 
\begin{itemize}
    \item Conditional image generation: 50K ImageNet~\cite{deng2009imagenet}-256$\times$256 samples generated by DiT~\cite{peebles2023scalable}.  
    \item Text-to-image generation: 5K COCO-prompt~\cite{lin2014microsoft} samples generated by Stable Diffusion v3.5 (d=38).
\end{itemize}

\noindent\textbf{Metrics.}  
For conditional generation, we report Fréchet Inception Distance (FID), Inception Score (IS), and improved Precision/Recall~\cite{kynkaanniemi2019improved}.  
For text-to-image generation, we use CLIP score~\cite{radford2021learning}, ImageReward (IR)~\cite{xu2023imagereward}, and an LLM-based evaluation for text–image alignment and perceptual quality.

\noindent\textbf{Protocol.}  
All experiments vary the number of function evaluations (NFE=$5\!-\!10$) with fixed random seeds for fair comparison, focusing on the low-NFE regime most relevant to real-time deployment.  
Experiments are conducted on a workstation equipped with 8$\times$NVIDIA A100 GPUs (40GB each). The code is available at \url{https://github.com/jinc7461/AFloPS}.

\subsection{Main Results on DiT Backbone}

We compare A-FloPS with representative training-free samplers: DDIM~\cite{song2021denoising}, DPM-Solver++~\cite{lu2022dpm}, and UniPC~\cite{zhao2023unipc}.  
Following~\cite{peebles2023scalable}, we use a CFG scale of $1.5$.  
For stability, $\lambda^{(n)}$ is constrained to $[-1,1]$, and $a,b$ in~\eqref{eq:ab} retain terms only up to second order in $\Delta t$.

Table~\ref{tab:main_dit} shows that A-FloPS consistently outperforms all baselines.  
At $\text{NFE}=5$, it achieves a notably lower FID than the next-best method, confirming its efficiency in the few-step regime.  
The advantage persists as NFE increases, demonstrating the robustness of our trajectory reparameterization and adaptive integration.

Qualitative results in Figure~\ref{fig:qualitative} further support these findings: competing methods tend to yield overly smooth or blurry edges at $\text{NFE}=5$, while A-FloPS produces sharper textures and more coherent structures.  
These results align with our objective of improving temporal regularity in the sampling trajectory—particularly critical for few-step generation.  
Additional examples are provided in the appendix.

\begin{figure}[!ht]
\centering
\begin{subfigure}[b]{0.9\columnwidth}
    \centering
    \includegraphics[width=0.88\linewidth]{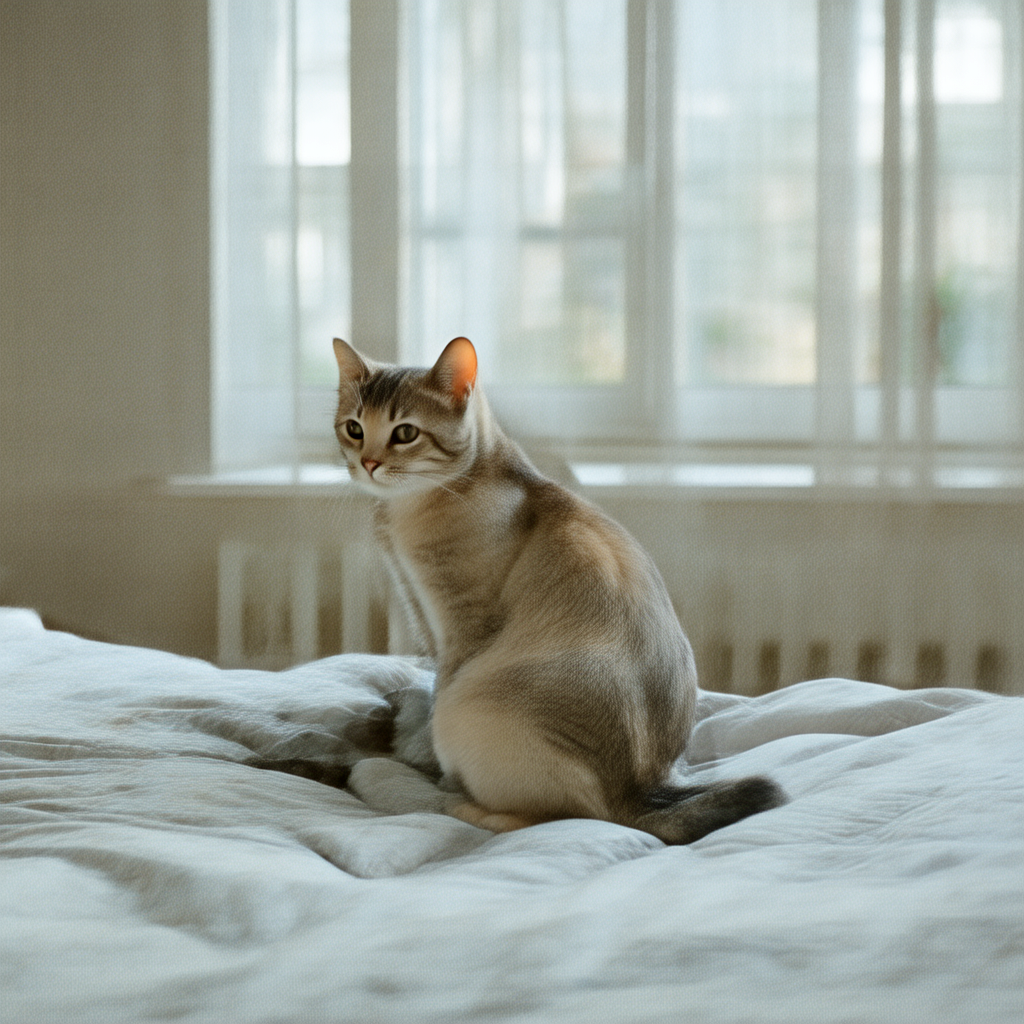}
    \caption{A-Euler (Ours)}
    \label{fig:a-euler}
\end{subfigure}

\begin{subfigure}[b]{0.9\columnwidth}
    \centering
    \includegraphics[width=0.88\linewidth]{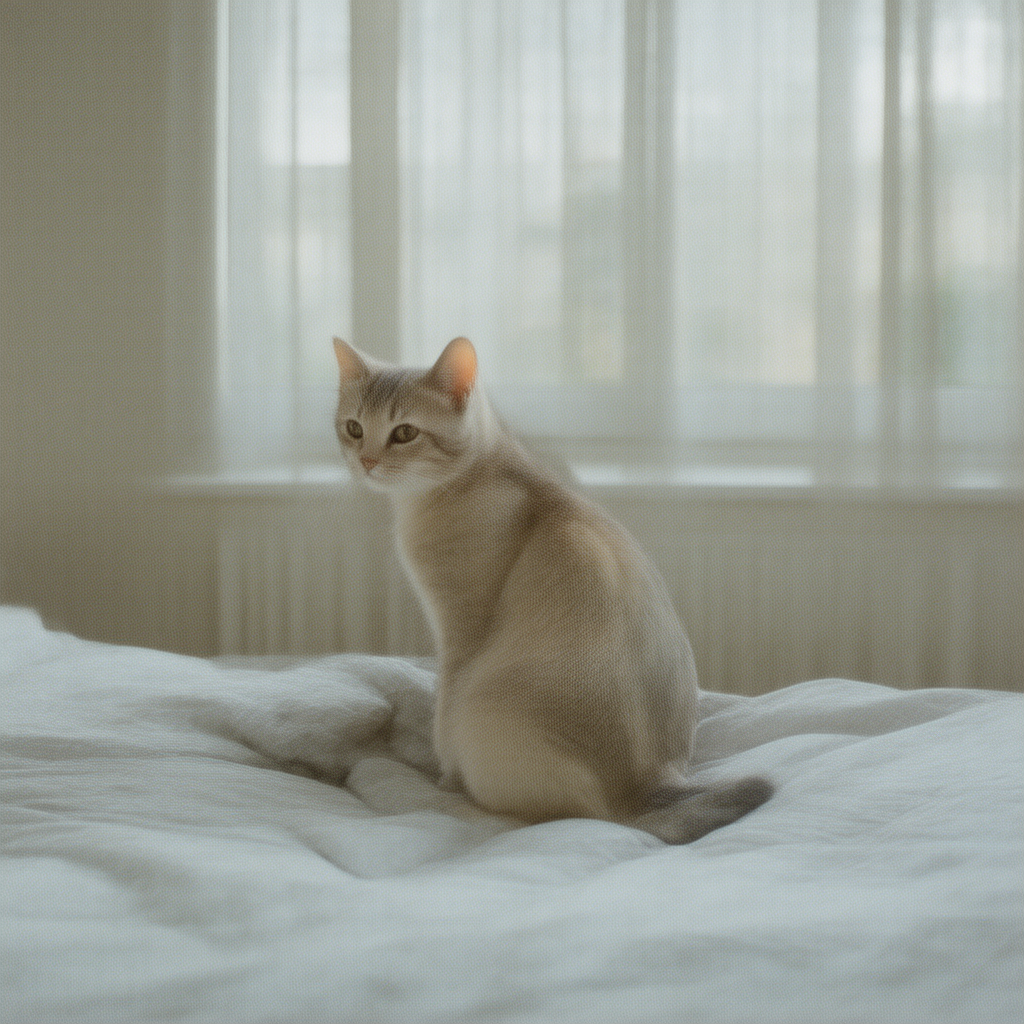}
    \caption{Euler (Baseline)}
    \label{fig:euler}
\end{subfigure}
\caption{
Qualitative comparison between A-Euler and baseline Euler on SDv3.5 (d=38) at $\text{NFE}=5$.  
Prompt: \textit{``A cat is sitting on top of a bed in front of a living room''}.  
A-Euler produces sharper textures and more coherent structures.}
\label{fig:aeuler_vs_euler}
\end{figure}

\subsection{Ablation Study}
\paragraph{Flow Path Reparameterization without Adaptive Mechanism.}
To isolate the contribution of flow-path reparameterization, we remove the adaptive mechanism while retaining the flow-transformed velocity field.  
As shown in Table~\ref{tab:ablation_flow_only}, this non-adaptive variant (FloPS) almost consistently surpasses DDIM across all NFEs, demonstrating that reparameterizing the sampling trajectory alone yields a more sample‑efficient process and achieves better alignment with the target distribution. 
While it falls short of the perceptual quality attained by the full A‑FloPS, it still delivers markedly sharper and more coherent results than DDIM, particularly in the challenging low‑NFE regime.  
This finding suggests that re‑normalizing the sampling path of existing diffusion models offers a practical and virtually cost‑free improvement.

\begin{table}[ht]
\centering
\small
\begin{tabular}{c|ccc|ccc}
\toprule
\multirow{2}{*}{NFE} & \multicolumn{3}{c|}{Euler} & \multicolumn{3}{c}{A-Euler (Ours)} \\
\cmidrule(lr){2-4} \cmidrule(lr){5-7}
 & CLIP$\uparrow$ & IR$\uparrow$ & GPT$\uparrow$ & CLIP$\uparrow$ & IR$\uparrow$ & GPT$\uparrow$ \\
\midrule
5 & 0.316 & 0.300 & 0.172 & \textbf{0.319} & \textbf{0.579} & \textbf{0.828} \\
7 & \textbf{0.319} & 0.629 & 0.234 & 0.318 & \textbf{0.785} & \textbf{0.766} \\
9 & \textbf{0.319} & 0.755 & 0.315 & 0.317 & \textbf{0.843} & \textbf{0.685} \\
\bottomrule
\end{tabular}
\caption{
Comparison of Euler and A-Euler on SDv3.5 under different NFEs ($\text{CFG}=2.0$).  
Higher CLIP, IR, and GPT indicate better performance.
}
\label{tab:ablation_sdv3}
\end{table}

\paragraph{Effectiveness of Adaptivity in Flow-Based Models.} 
Building on the ablation results above, where A‑FloPS consistently outperforms its non-adaptive counterpart (FloPS), it is natural to ask: while FloPS reparameterizes a diffusion model into a flow‑matching (FM) model and the adaptive mechanism is tailored to FM dynamics, can this mechanism also benefit \emph{native} FM models that are not obtained via reparameterization?

To answer this, we integrate the adaptive mechanism into a representative FM-based model, Stable Diffusion v3.5 (d=38).  
As shown in Table~\ref{tab:ablation_sdv3}, it yields consistent gains across most metrics, with the largest improvements at low NFEs.  
Although CLIP scores for A‑Euler are slightly lower than Euler at $N=7,9$, this is likely due to CLIP’s known saturation and stylistic bias~\cite{yarom2023seetrue}.  
In contrast, IR and GPT-based metrics—better aligned with human judgment—show substantial improvements. The GPT-based evaluation compares two images from the same prompt, selecting the one with better text alignment and perceptual quality; details are in the appendix.
Figure~\ref{fig:aeuler_vs_euler} further shows that A‑Euler produces sharper textures and more coherent structures than baseline Euler at $\text{NFE}=5$. 

These improvements on a native FM model suggest that the adaptive mechanism is not tied to the specifics of our diffusion-to-flow reparameterization.  
In fact, the mechanism is agnostic to how the FM velocity field is obtained—whether learned directly, distilled from a diffusion model, or analytically derived.  
It can thus be applied to a broad range of deterministic generative ODEs, such as rectified flows~\cite{liu2022flow} and hybrid diffusion–flow systems.  
While our experiments focus on image generation, the method naturally extends to other modalities such as audio or video synthesis.

Our experiments lead to three main observations. First, flow-based reparameterization substantially improves sampling efficiency by producing trajectories that are inherently more amenable to numerical integration. Second, the adaptive mechanism provides additional accuracy gains, particularly in the low-NFE regime. Finally, combining these two components allows A-FloPS to achieve both high sampling speed and strong generation quality, all without requiring model retraining.

\section{Conclusion}
We introduced \textbf{A-FloPS}, a training-free framework that accelerates diffusion sampling by reparameterizing arbitrary sampling paths into flow-matching (FM) trajectories and augmenting them with an adaptive velocity decomposition. This addresses two core limitations of prior training-free methods: (i) inefficient integration of conventional diffusion trajectories, and (ii) reduced benefits of high-order solvers under weakly time-varying velocity fields.

The reparameterization alone yields integration-friendly paths, while the adaptive mechanism restores the accuracy advantages of high-order integration in the few-step regime. Together, they deliver superior generation quality and semantic fidelity with far fewer function evaluations, across both diffusion and native FM models.

Experiments on ImageNet and COCO show consistent state-of-the-art performance among training-free samplers, especially at extremely low NFEs. Future work will extend A-FloPS to multi-modal synthesis, real-time generation, and richer temporal models for the adaptive coefficient.
\section{Ethics Statement}
This work focuses on improving the sampling efficiency of pretrained diffusion models.
All experiments are conducted on publicly available datasets with appropriate licenses (ImageNet, COCO).
No personally identifiable information or sensitive data is used.
The proposed method does not introduce new ethical risks beyond those already associated with generative models, such as potential misuse for generating misleading content.
We emphasize that our method should be applied responsibly and strictly for research and non-harmful purposes.
No human subjects or animals were involved, and no IRB approval was required.
\section{Acknowledgements}
This work was supported by NSAF (Grant No. U2230201). All authors are affiliated with the Department of Electronic Engineering, Tsinghua University. In addition, Y. Gu is also affiliated with the Tsinghua Shenzhen International Graduate School and the Beijing National Research Center for Information Science and Technology at Tsinghua University.

\bibliography{aaai2026}

\onecolumn
\appendix
\section{Proof of Theorem.\ref{thm:score-to-velocity}}

The core idea of the proof is to first align the noisy variables of the flow matching and diffusion models via a linear rescaling, and then relate the flow matching velocity field to the diffusion model's score function through the conditional expectation $\mathbb{E}\!\left[\x_{\text{data}} \mid \x_{\text{noisy}}\right]$. The velocity field is obtained from Eq.~\eqref{eq:vstar}, while the score function follows from Tweedie’s formula.

We start from Eq.~\eqref{eq:vstar}:
\begin{align}
\label{eq:app:1}
    \vv^*(\x_t, t) &= \mathbb{E}_q\!\left[\x_1 - \x_0 \,\middle|\, \x_t \right] \notag\\
    &=\mathbb{E}_q\!\left[\x_1 - \x_0 \,\middle|\, (1 - t) \x_0 + t \x_1=\x_t \right] \notag\\
    &=\mathbb{E}_q\!\left[\x_1 - \frac{\x_t-t \x_1}{1 - t} \,\middle|\, (1 - t) \x_0 + t \x_1=\x_t \right] \notag\\
    &=\mathbb{E}_q\!\left[\frac{\x_1 - \x_t}{1 - t} \,\middle|\, \x_t \right] \notag\\
    &=\frac{1}{1 - t}\mathbb{E}_q\!\left[\x_1  \,\middle|\, \x_t \right]-\frac{1}{1 - t}\x_t \notag\\
    &=\frac{1}{1 - t}\int \x_1 \frac{q_1(\x_1)q_{t|1}(\x_t\mid \x_1)}{q_t(\x_t)}\dd \x_1-\frac{1}{1 - t}\x_t \notag\\
    &=\frac{1}{1 - t}\int \x_1 \frac{\Gauss(\x_t;t\x_1,(1-t)^2\eye)}{q_t(\x_t)}q_1(\x_1)\dd \x_1-\frac{1}{1 - t}\x_t.
\end{align}

For the diffusion model, Tweedie’s formula states:
\begin{align}
\label{eq:app:2}
    \frac{1}{\bar{\alpha}_\tau}\left[\x_{\tau}+\sigma_\tau^2\nabla\log p_\tau(\x_\tau)\right]&=\mathbb{E}_p\!\left[\x_0|\, \x_\tau \right] \notag\\
    &=\int \x_0 \frac{p_0(\x_0)p_{\tau\mid 0}(\x_\tau|\x_0)}{p_\tau(\x_\tau)}p_0(\x_0)\dd \x_0 \notag\\
    &=\int \x_0 \frac{\Gauss(\x_\tau;\bar{\alpha}_\tau\x_0,\sigma_\tau^2\eye)}{p_\tau(\x_\tau)}p_0(\x_0)\dd \x_0.
\end{align}

With $t = \frac{1}{1 + \sigma_\tau / \bar{\alpha}_\tau}$, the Gaussian term in the flow matching model can be rewritten as:
\begin{align}
    \Gauss(\x_t;t\x_1,(1-t)^2\eye)&=\frac{1}{(2\pi)^{d/2} (1-t)^d}
\exp\left(
 -\frac{\|\x_t - t\x_1\|^2}{2(1-t)^2}
\right)  \notag\\
&=\frac{\left(\bar{\alpha}_\tau + \sigma_\tau\right)^d}{(2\pi)^{d/2} \sigma_\tau^d}
\exp\left( -\frac{\left(\bar{\alpha}_\tau + \sigma_\tau\right)^2 \left\|\x_t - \frac{\bar{\alpha}_\tau}{\bar{\alpha}_\tau + \sigma_\tau}\x_1 \right\|^2}{2\sigma_\tau^2} \right)\notag\\
&=\frac{\left(\bar{\alpha}_\tau + \sigma_\tau\right)^d}{(2\pi)^{d/2} \sigma_\tau^d}
\exp\left( -\frac{ \left\|\left(\bar{\alpha}_\tau + \sigma_\tau\right)\x_t - {\bar{\alpha}_\tau}\x_1 \right\|^2}{2\sigma_\tau^2} \right),
\end{align}
and the marginal pdf can be written as
\begin{align}
    q_t(\x_t)&=\int q_1(\x_1)q_{t\mid 1}(\x_t\mid \x_1)\dd \x_1 \notag\\
    &=\int q_1(\x_1)\Gauss(\x_t;t\x_1,(1-t)^2\eye)\dd \x_1 \notag\\
    &=\int q_1(\x_1)\frac{\left(\bar{\alpha}_\tau + \sigma_\tau\right)^d}{(2\pi)^{d/2} \sigma_\tau^d}
\exp\left( -\frac{ \left\|\left(\bar{\alpha}_\tau + \sigma_\tau\right)\x_t - {\bar{\alpha}_\tau}\x_1 \right\|^2}{2\sigma_\tau^2} \right)\dd \x_1.
\end{align}

Therefore, we have
\begin{align}
    \frac{\Gauss(\x_t;t\x_1,(1-t)^2\eye)}{q_t(\x_t)}=
    \frac{\exp\left( -\frac{ \left\|\left(\bar{\alpha}_\tau + \sigma_\tau\right)\x_t - {\bar{\alpha}_\tau}\x_1 \right\|^2}{2\sigma_\tau^2} \right)}
    {\int q_1(\x_1)\exp\left( -\frac{ \left\|\left(\bar{\alpha}_\tau + \sigma_\tau\right)\x_t - {\bar{\alpha}_\tau}\x_1 \right\|^2}{2\sigma_\tau^2} \right)\dd \x_1}.
\end{align}

Similarly, we have 
\begin{align}
    \frac{\Gauss(\x_\tau;\bar{\alpha}_\tau\x_0,\sigma_\tau^2\eye)}{p_\tau(\x_\tau)}=
    \frac{\exp\left( -\frac{ \left\|\left(\bar{\alpha}_\tau + \sigma_\tau\right)\x_t - {\bar{\alpha}_\tau}\x_0 \right\|^2}{2\sigma_\tau^2} \right)}
    {\int p_0(\x_0)\exp\left( -\frac{ \left\|\left(\bar{\alpha}_\tau + \sigma_\tau\right)\x_t - {\bar{\alpha}_\tau}\x_0 \right\|^2}{2\sigma_\tau^2} \right)\dd \x_0}.
\end{align}

Then, substituting \(\x_\tau = (\bar{\alpha}_\tau + \sigma_\tau)\) into~\eqref{eq:app:2}, we obtain
\begin{align}
\label{eq:app:3}
    \frac{1}{\bar{\alpha}_\tau}\left[(\bar{\alpha}_\tau+\sigma_\tau)\x_{t}+\sigma_\tau^2\nabla\log p_\tau((\bar{\alpha}_\tau+\sigma_\tau)\x_t)\right]&=\int \x_0 \frac{\exp\left( -\frac{ \left\|\left(\bar{\alpha}_\tau + \sigma_\tau\right)\x_t - {\bar{\alpha}_\tau}\x_0 \right\|^2}{2\sigma_\tau^2} \right)}
    {\int p_0(\x_0)\exp\left( -\frac{ \left\|\left(\bar{\alpha}_\tau + \sigma_\tau\right)\x_t - {\bar{\alpha}_\tau}\x_0 \right\|^2}{2\sigma_\tau^2} \right)\dd \x_0}p_0(\x_0)\dd \x_0 \notag \\
    &=\int \x_0 \frac{\Gauss(\x_t;t\x_0,(1-t)^2\eye)}{q_t(\x_t)}p_0(\x_0)\dd \x_0.
\end{align}

Since $p_0 = q_1$ and relabeling $\x_0$ as $\x_1$, Eq.~\eqref{eq:app:3} becomes:
\begin{equation}
\label{eq:app:4}
    \frac{1}{\bar{\alpha}_\tau}\left[(\bar{\alpha}_\tau+\sigma_\tau)\x_{t}+\sigma_\tau^2\nabla\log p_\tau((\bar{\alpha}_\tau+\sigma_\tau)\x_t)\right]=\int \x_1 \frac{\Gauss(\x_t;t\x_1,(1-t)^2\eye)}{q_t(\x_t)}q_1(\x_1)\dd \x_1.
\end{equation}

Substitude~\eqref{eq:app:4} in~\eqref{eq:app:1}, we have
\begin{align}
    (1-t)\vv^*(\x_t, t)&=\int \x_1 \frac{\Gauss(\x_t;t\x_1,(1-t)^2\eye)}{q_t(\x_t)}q_1(\x_1)\dd \x_1-\x_t \notag\\
    &=\frac{1}{\bar{\alpha}_\tau}\left[(\bar{\alpha}_\tau+\sigma_\tau)\x_{t}+\sigma_\tau^2\nabla\log p_\tau((\bar{\alpha}_\tau+\sigma_\tau)\x_t)\right]-\x_t
    \notag\\
    &=\frac{1}{\bar{\alpha}_\tau}\left[\sigma_\tau\x_{t}+\sigma_\tau^2\nabla\log p_\tau((\bar{\alpha}_\tau+\sigma_\tau)\x_t)\right]
    \notag\\
    &=\frac{\sigma_\tau}{\bar{\alpha}_\tau}\left[\x_{t}+\sigma_\tau\nabla\log p_\tau((\bar{\alpha}_\tau+\sigma_\tau)\x_t)\right].
\end{align}

\newpage
\section{GPT-based Evaluation Details}
For the GPT-based evaluation in the main paper, we use a pairwise comparison protocol to assess both image quality and text–image alignment. This approach yields a more nuanced assessment than traditional metrics by directly comparing images generated from the same prompt with different sampling methods.

\subsection{Evaluation Protocol}
We employ GPT-4o (temperature = 0.0) to ensure deterministic outputs.  
For each prompt, the model receives two images—one from the baseline and one from our A-FloPS sampler—and makes a binary choice based on overall quality (visual authenticity + prompt adherence).

\subsection{System Prompt}
\begin{lstlisting}[caption={System prompt for GPT-4o evaluator}, 
                   label={lst:system-prompt}, 
                   basicstyle=\ttfamily\small, 
                   frame=single,
                   breaklines=true]
You are an image quality assessment assistant. 
You will receive two images with a prompt for comparison.
Task: Evaluate both images considering two aspects:
Image quality and authenticity
How well each image aligns with the given prompt

Rules:
- Return 0 if the first image is better overall
- Return 1 if the second image is better overall
- Must return ONLY 0 or 1
- Do not provide any explanation or additional text
- Do not include any punctuation marks
\end{lstlisting}

\subsection{User Prompt}

\begin{lstlisting}[caption={User prompt for GPT-4o evaluator}, 
                   label={lst:user-query}, 
                   basicstyle=\ttfamily\small, 
                   frame=single,
                   breaklines=true，
                   keepspaces=false]
Based on this prompt: '[prompt]', compare these two images. Consider both image quality/
authenticity and prompt alignment. Return 0 if the first image is better overall, return 1 
if the second is better. Return only the number 0 or 1.
Here is the first image:
'[Image1]'
Here is the second image:
'[Image2]'
Return only 0 or 1
\end{lstlisting}

\section{Additional Visualizations}
This section provides additional qualitative results to complement the quantitative analysis in the main body of the paper. We present visual comparisons on both the DiT model for conditional image generation and the Stable Diffusion v3.5 model for text-to-image synthesis. These examples serve to visually demonstrate the improvements in sample quality, detail fidelity, and structural coherence achieved by our proposed methods.
\subsection{Dit}
The following figure provides a side-by-side comparison of samples generated from the DiT-L/2 model on ImageNet-256x256. All images were generated with an NFE of 5, a regime where many samplers struggle to produce high-quality outputs. As illustrated, A-FloPS consistently generates images with sharper textures (e.g., fur, grass) and more coherent global structures compared to strong baselines like DDIM, DPM-Solver++, and UniPC, underscoring its effectiveness for few-step conditional generation.
\begin{figure}[H]
    \centering
    \begin{subfigure}[b]{0.45\textwidth}
        \centering
        \includegraphics[width=\textwidth]{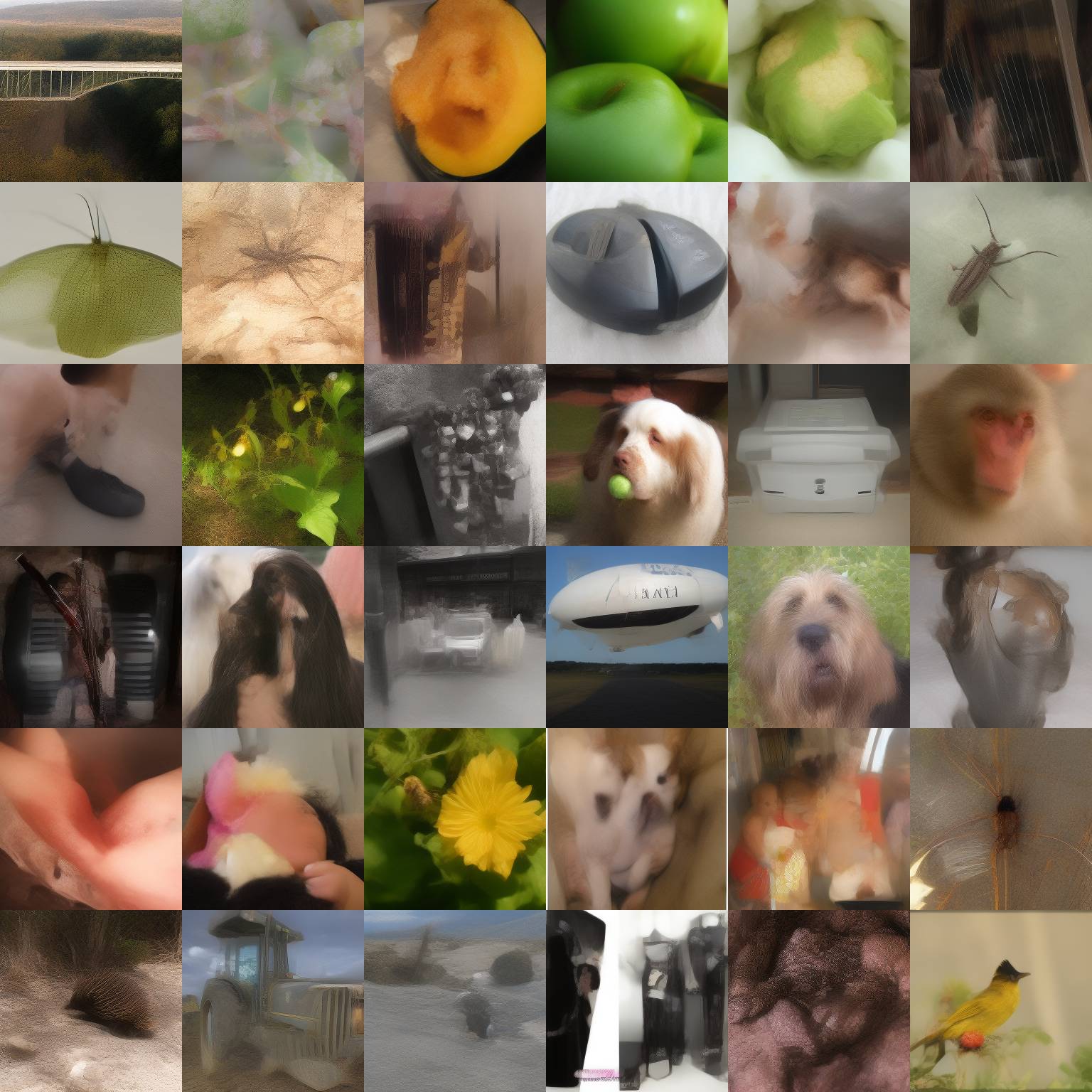}
        \caption{DDIM}
        \label{fig:subfig1}
    \end{subfigure}
    \hfill 
    \begin{subfigure}[b]{0.45\textwidth}
        \centering
        \includegraphics[width=\textwidth]{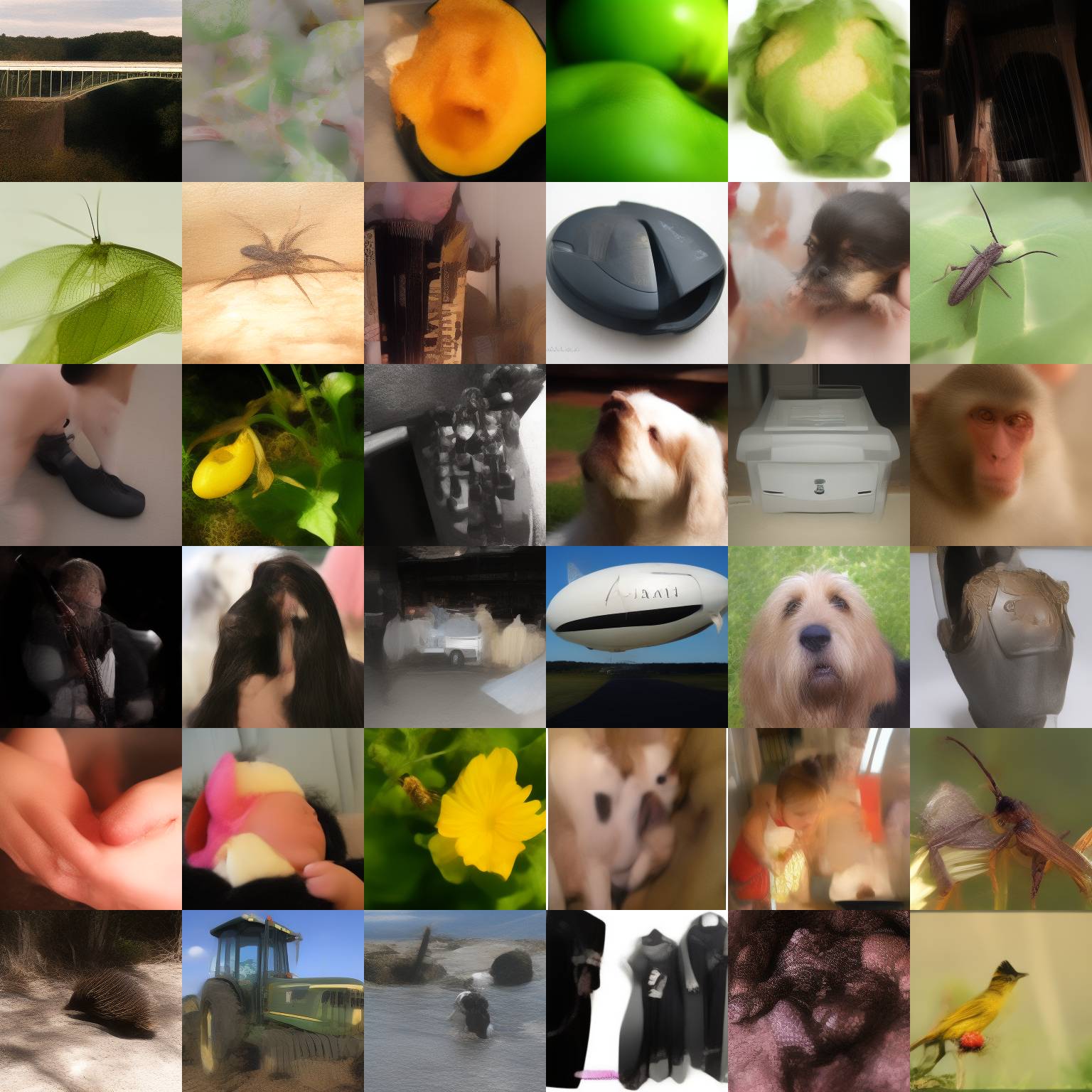}
        \caption{DPM-Solver++}
        \label{fig:subfig2}
    \end{subfigure}
    
    \vspace{0.5cm} 

    \begin{subfigure}[b]{0.45\textwidth}
        \centering
        \includegraphics[width=\textwidth]{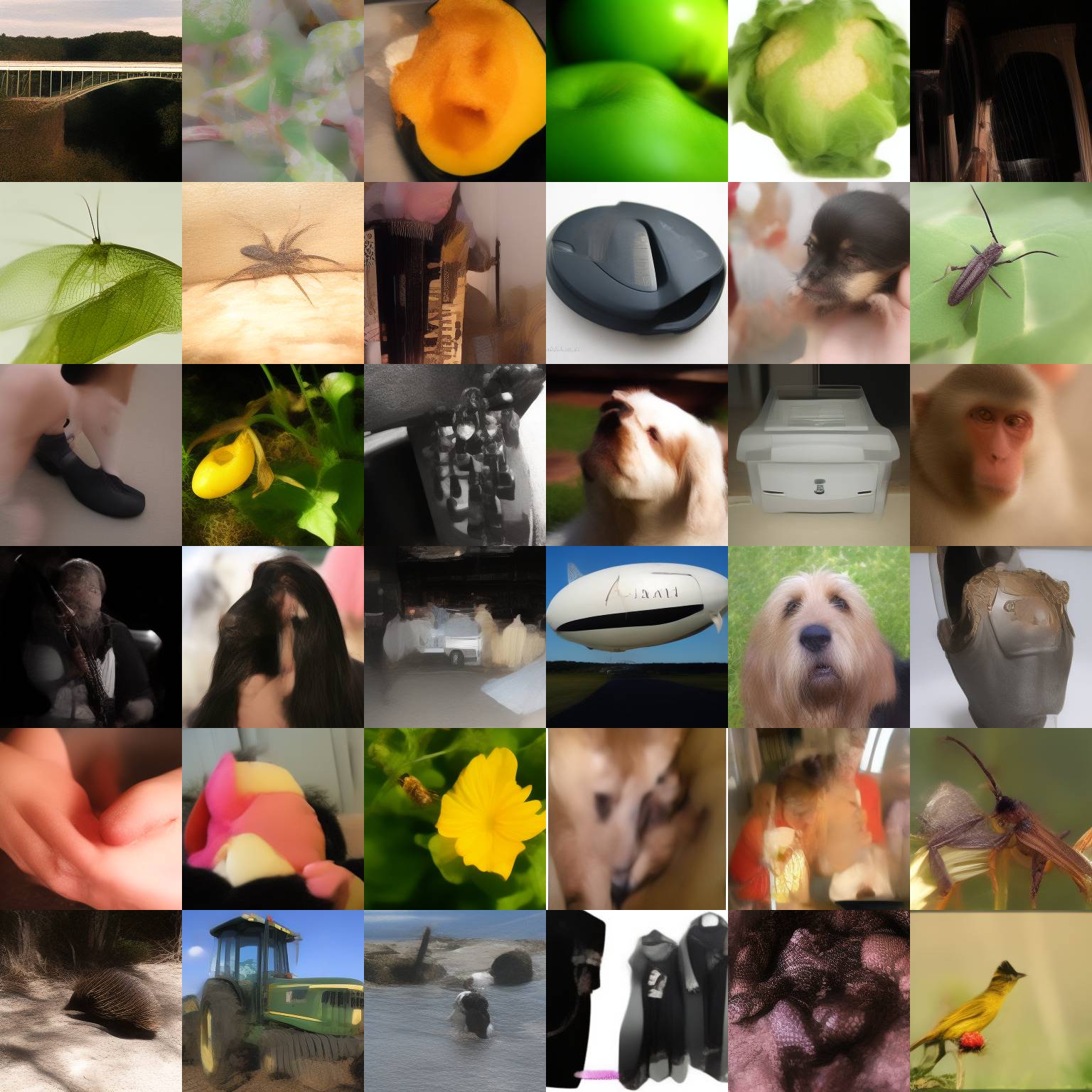}
        \caption{UniPC}
        \label{fig:subfig3}
    \end{subfigure}
    \hfill 
    \begin{subfigure}[b]{0.45\textwidth}
        \centering
        \includegraphics[width=\textwidth]{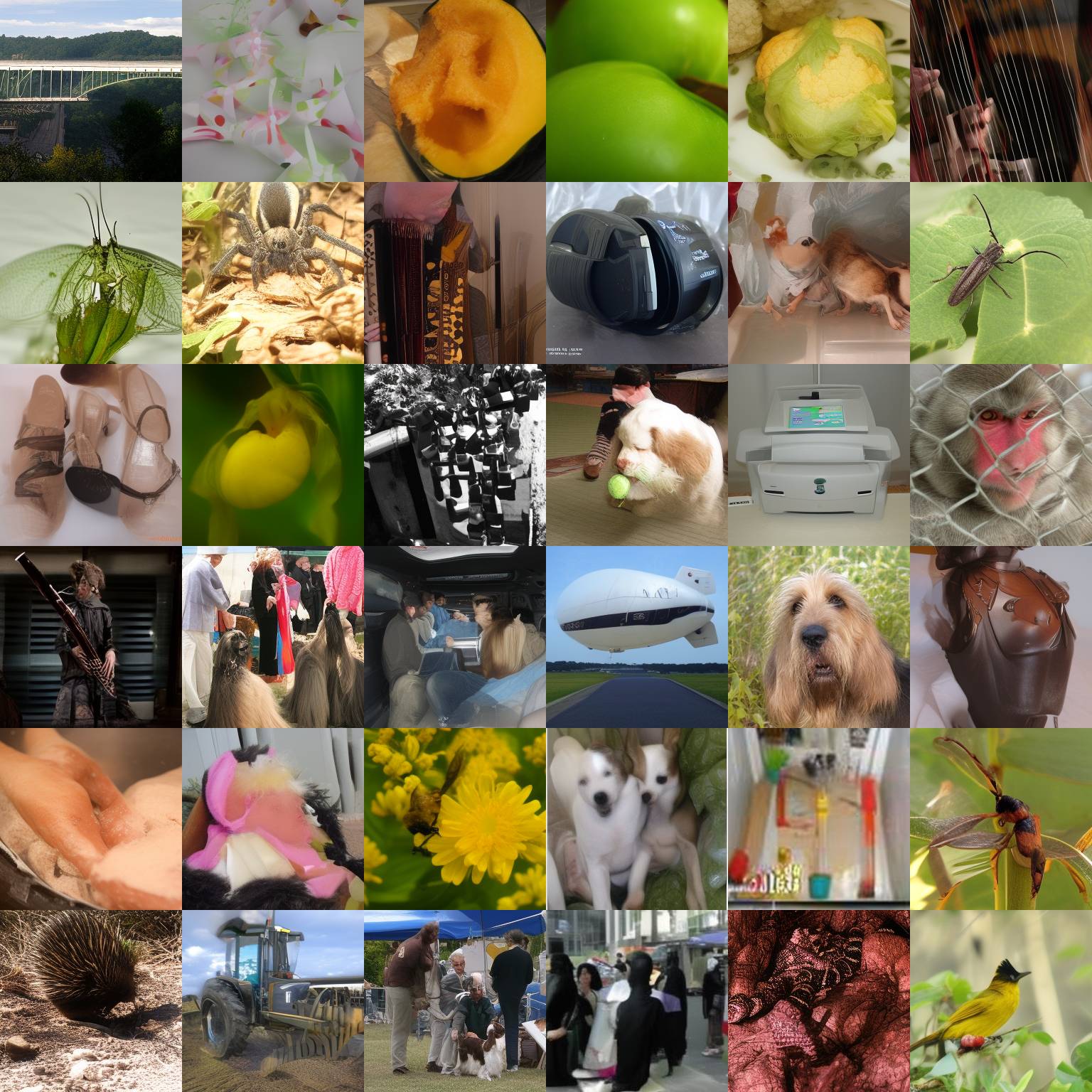} 
        \caption{A-FloPS}
        \label{fig:subfig4}
    \end{subfigure}
    
    \caption{Qualitative comparison of generated samples from different samplers on the DIT model with $\text{NFE}=5$. Our proposed A-FloPS (d) produces images with clearer structures, richer details compared to DDIM (a), DPM-Solver++ (b), and UniPC (c).
}
    \label{fig:four_subfigures}
\end{figure}

\subsection{Stable Diffusion}
To demonstrate the generalizability of our adaptive mechanism, we apply it to the native flow-based sampler of Stable Diffusion v3.5 (d=38), which we term 'A-Euler'. The figures below show paired comparisons between the standard Euler sampler and A-Euler for various text prompts at low NFEs. These examples highlight that the adaptive component significantly enhances perceptual quality, leading to more realistic details, fewer artifacts, and better overall composition, even without the full FloPS reparameterization.
\begin{figure}[H]
    \centering
    \begin{subfigure}[b]{0.32\textwidth}
        \centering
        \includegraphics[width=\textwidth]{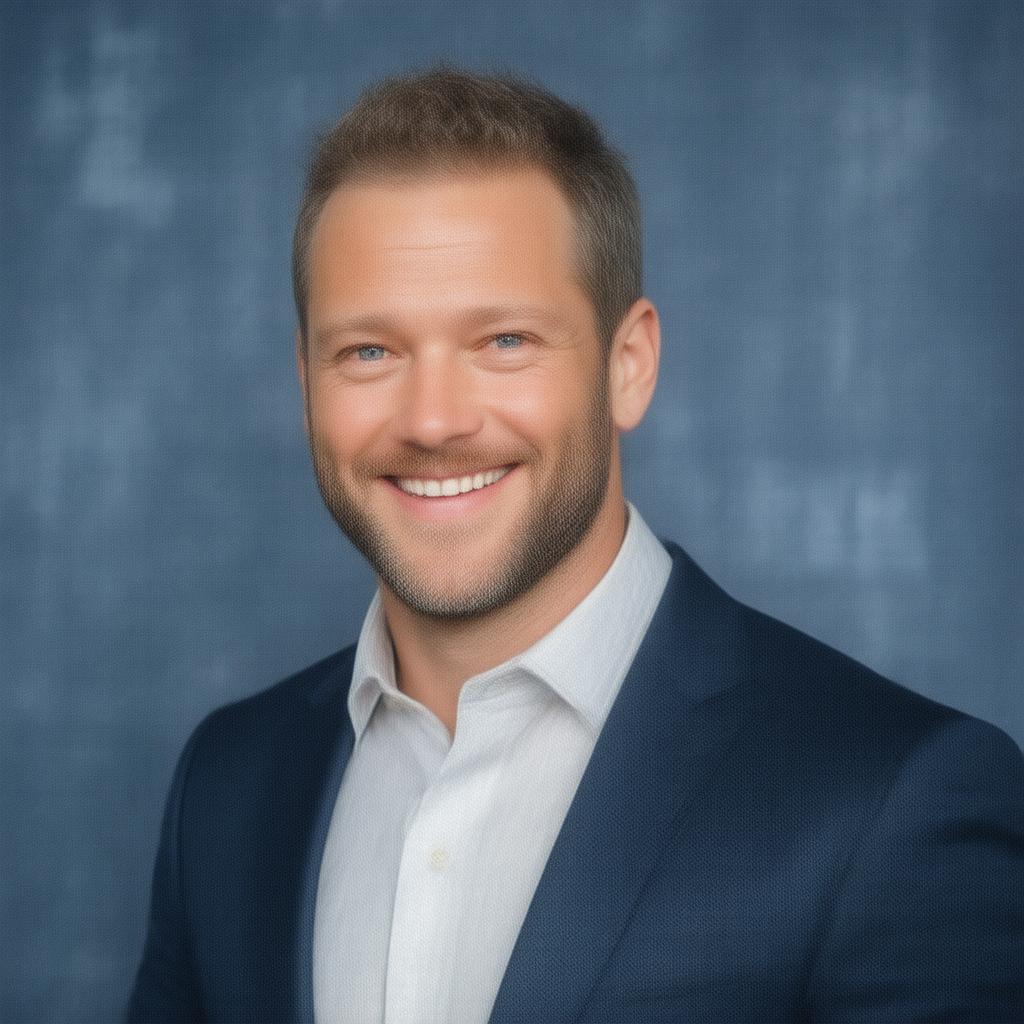}
        \caption*{Euler}
    \end{subfigure}
    \hspace{0.02\textwidth}
    \begin{subfigure}[b]{0.32\textwidth}
        \centering
        \includegraphics[width=\textwidth]{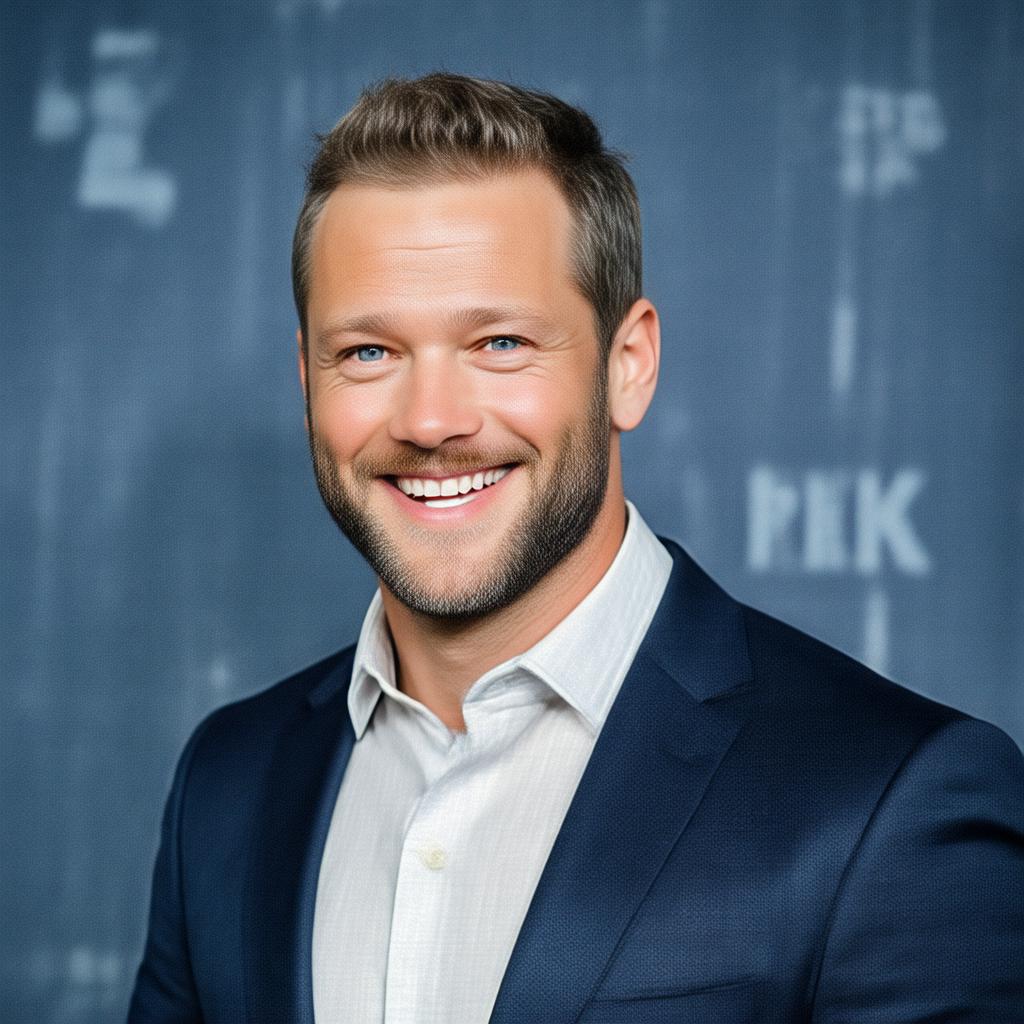}
        \caption*{A-Euler}
    \end{subfigure}
    \caption{Prompt: A man in a suit smiling at the camera}
    
    \begin{subfigure}[b]{0.32\textwidth}
        \centering
        \includegraphics[width=\textwidth]{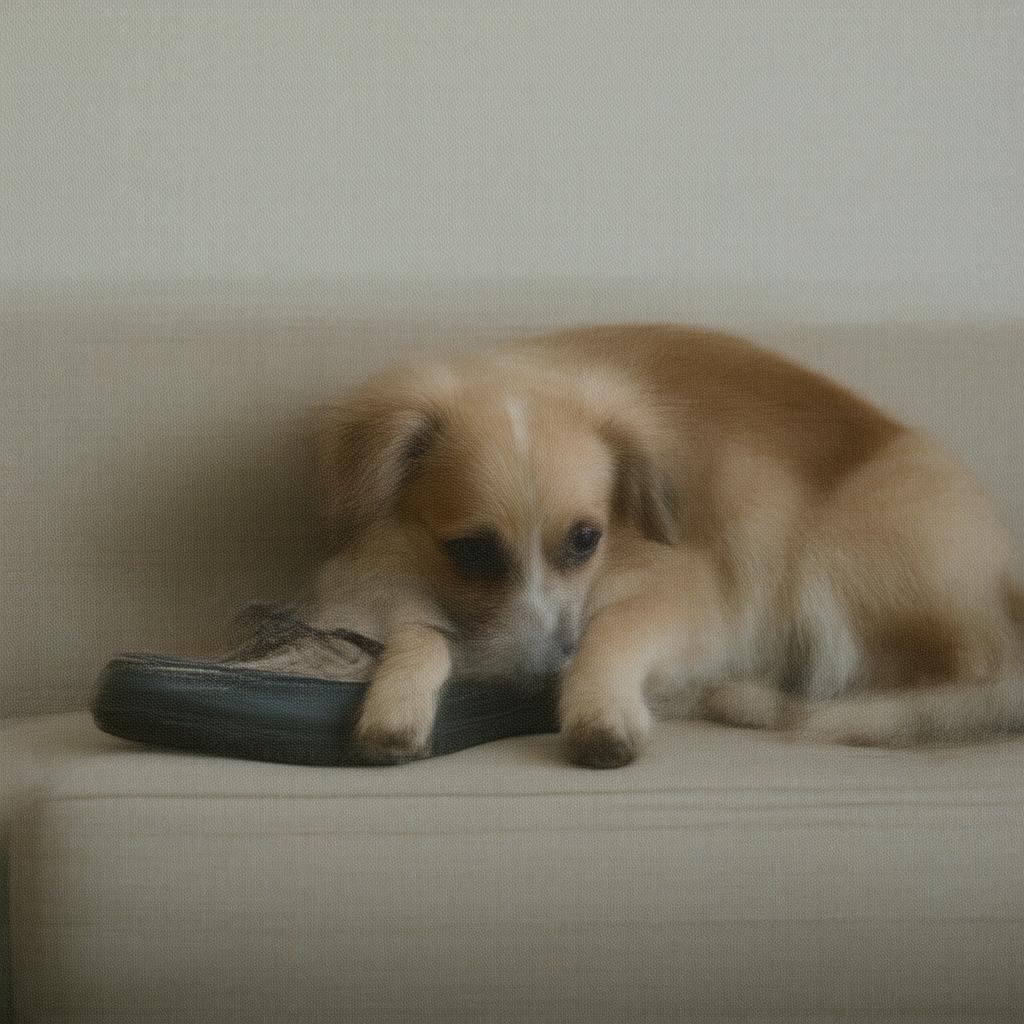}
        \caption*{Euler}
    \end{subfigure}
    \hspace{0.02\textwidth}
    \begin{subfigure}[b]{0.32\textwidth}
        \centering
        \includegraphics[width=\textwidth]{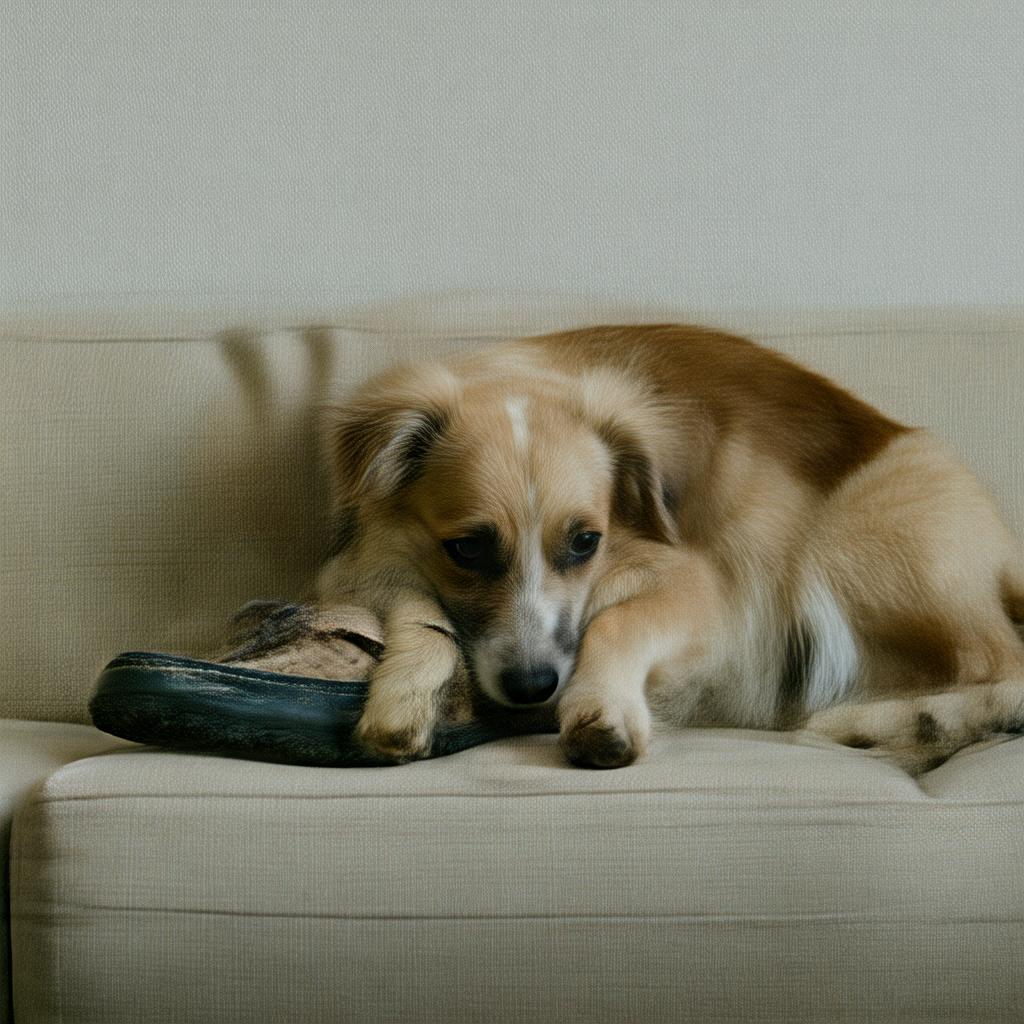}
        \caption*{A-Euler}
    \end{subfigure}
    \caption{Prompt: A small dog is laying on a couch with a shoe}
    
    \begin{subfigure}[b]{0.32\textwidth}
        \centering
        \includegraphics[width=\textwidth]{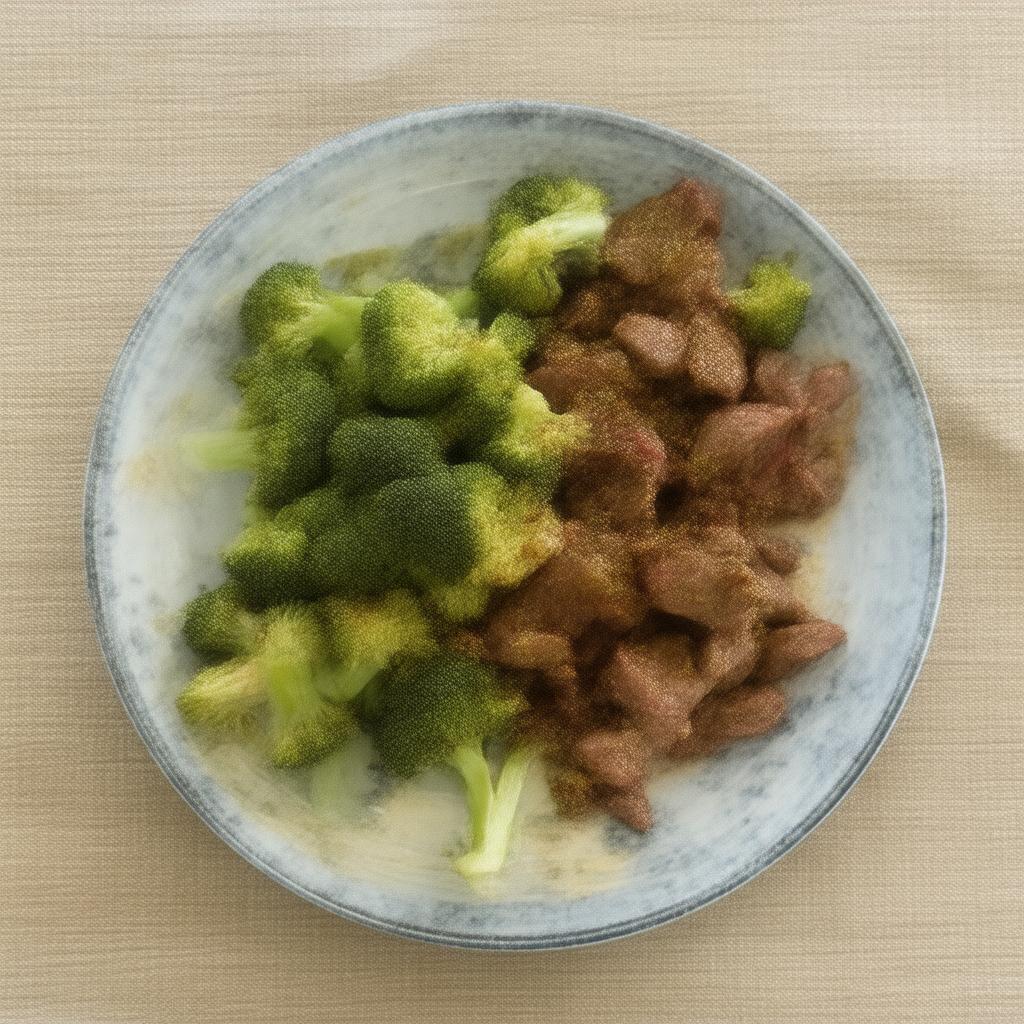}
        \caption*{Euler}
    \end{subfigure}
    \hspace{0.02\textwidth}
    \begin{subfigure}[b]{0.32\textwidth}
        \centering
        \includegraphics[width=\textwidth]{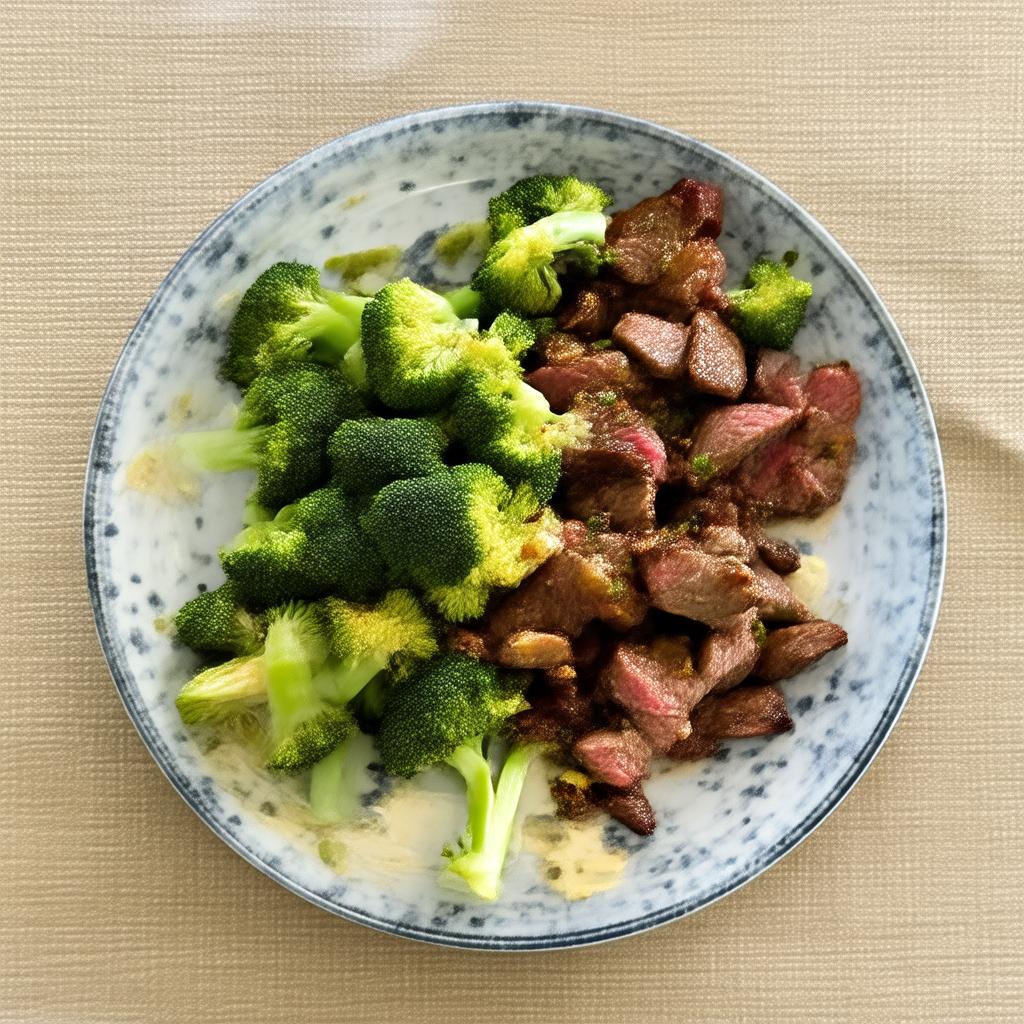}
        \caption*{A-Euler}
    \end{subfigure}
    \caption{Prompt: A plate of broccoli and meat on a table}
    \label{fig:paired_comparisons_1}
\end{figure}

\begin{figure}[H]
    \centering
    \begin{subfigure}[b]{0.32\textwidth}
        \centering
        \includegraphics[width=\textwidth]{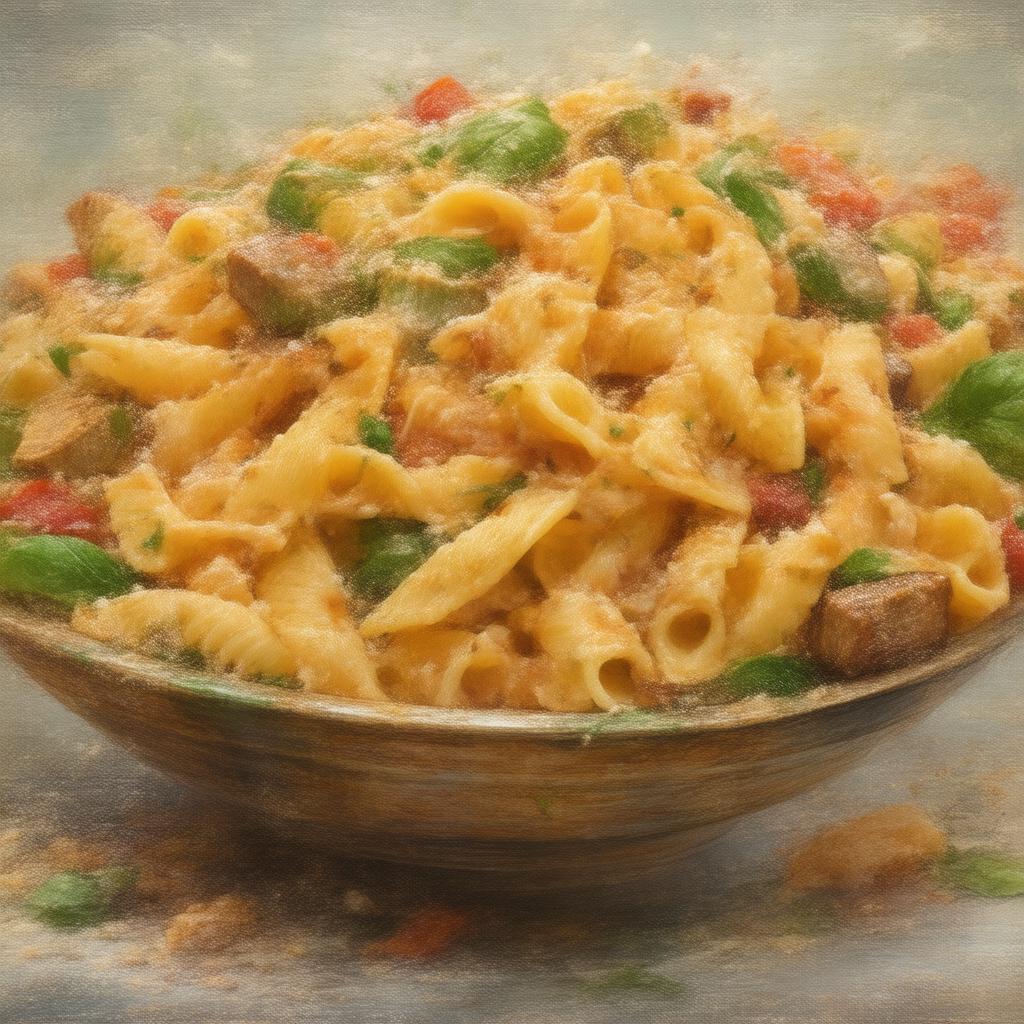}
        \caption*{Euler}
    \end{subfigure}
    \hspace{0.02\textwidth}
    \begin{subfigure}[b]{0.32\textwidth}
        \centering
        \includegraphics[width=\textwidth]{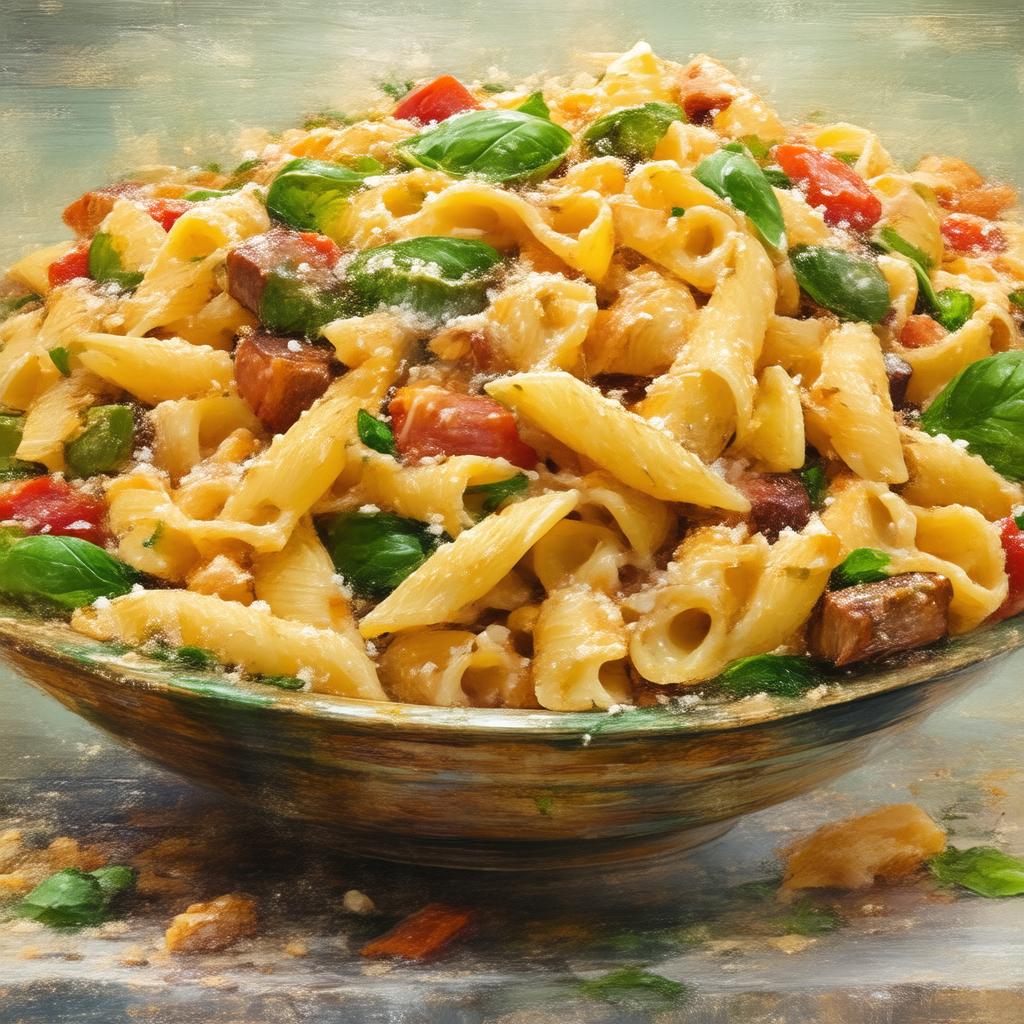}
        \caption*{A-Euler}
    \end{subfigure}
    \caption{Prompt: a large bowl full of pasta with many other foods}
    
    \begin{subfigure}[b]{0.32\textwidth}
        \centering
        \includegraphics[width=\textwidth]{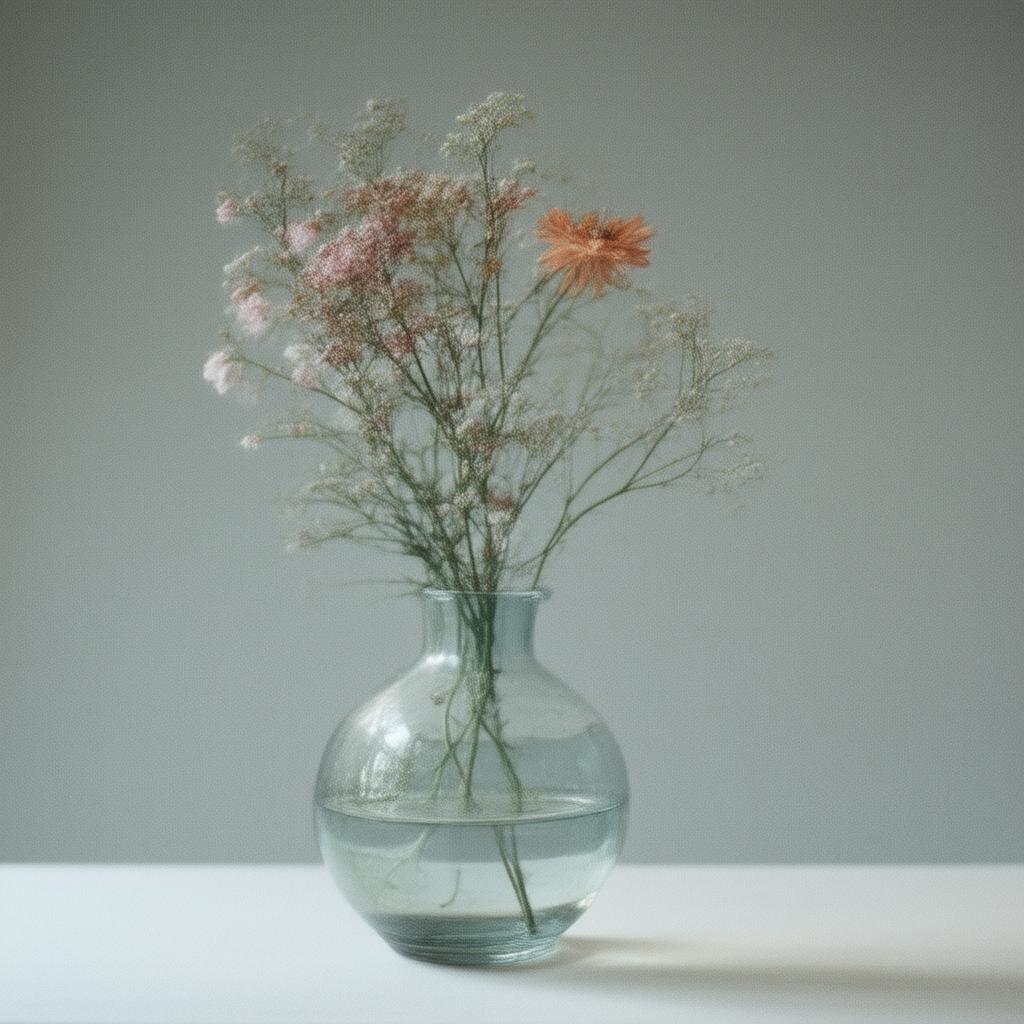}
        \caption*{Euler}
    \end{subfigure}
    \hspace{0.02\textwidth}
    \begin{subfigure}[b]{0.32\textwidth}
        \centering
        \includegraphics[width=\textwidth]{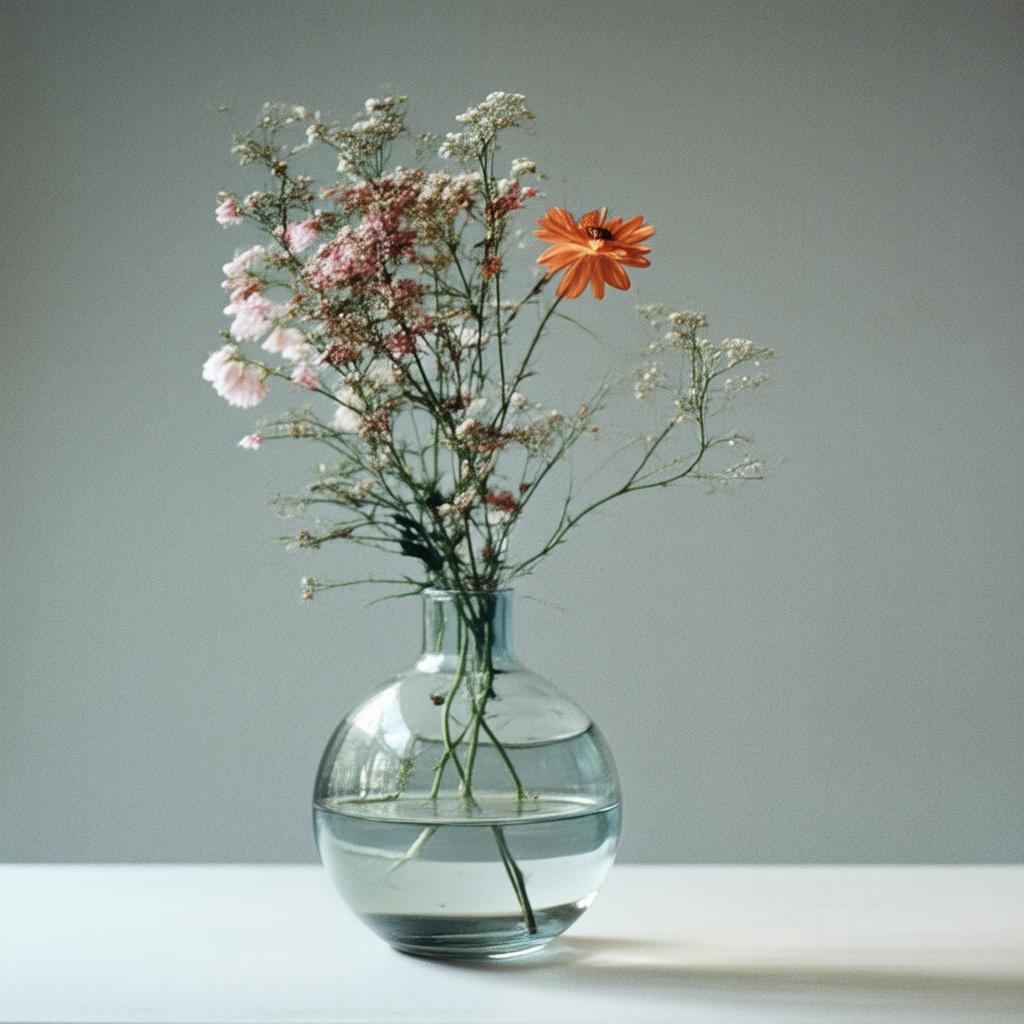}
        \caption*{A-Euler}
    \end{subfigure}
    \caption{Prompt: a glass vase with some flowers coming out of it}
    \label{fig:paired_comparisons_2}

    \begin{subfigure}[b]{0.32\textwidth}
        \centering
        \includegraphics[width=\textwidth]{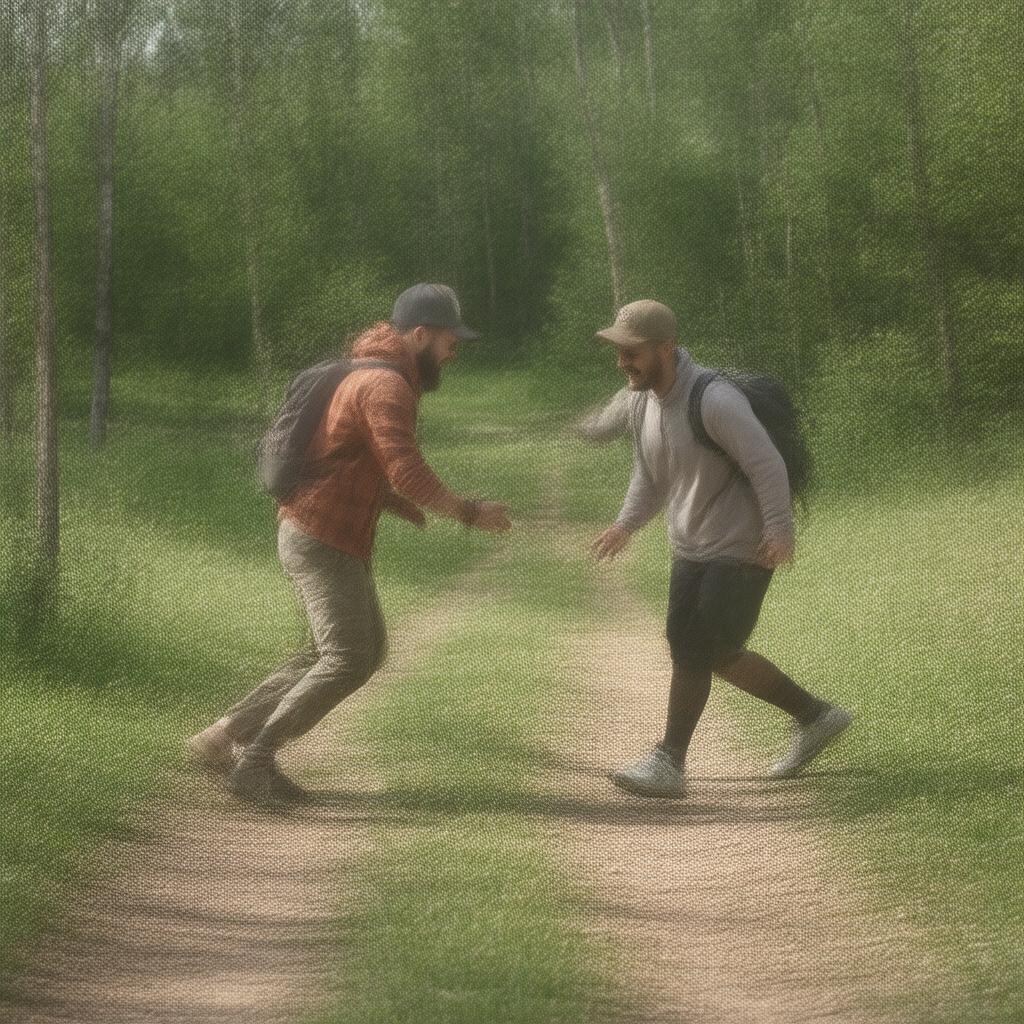}
        \caption*{Euler}
    \end{subfigure}
    \hspace{0.02\textwidth}
    \begin{subfigure}[b]{0.32\textwidth}
        \centering
        \includegraphics[width=\textwidth]{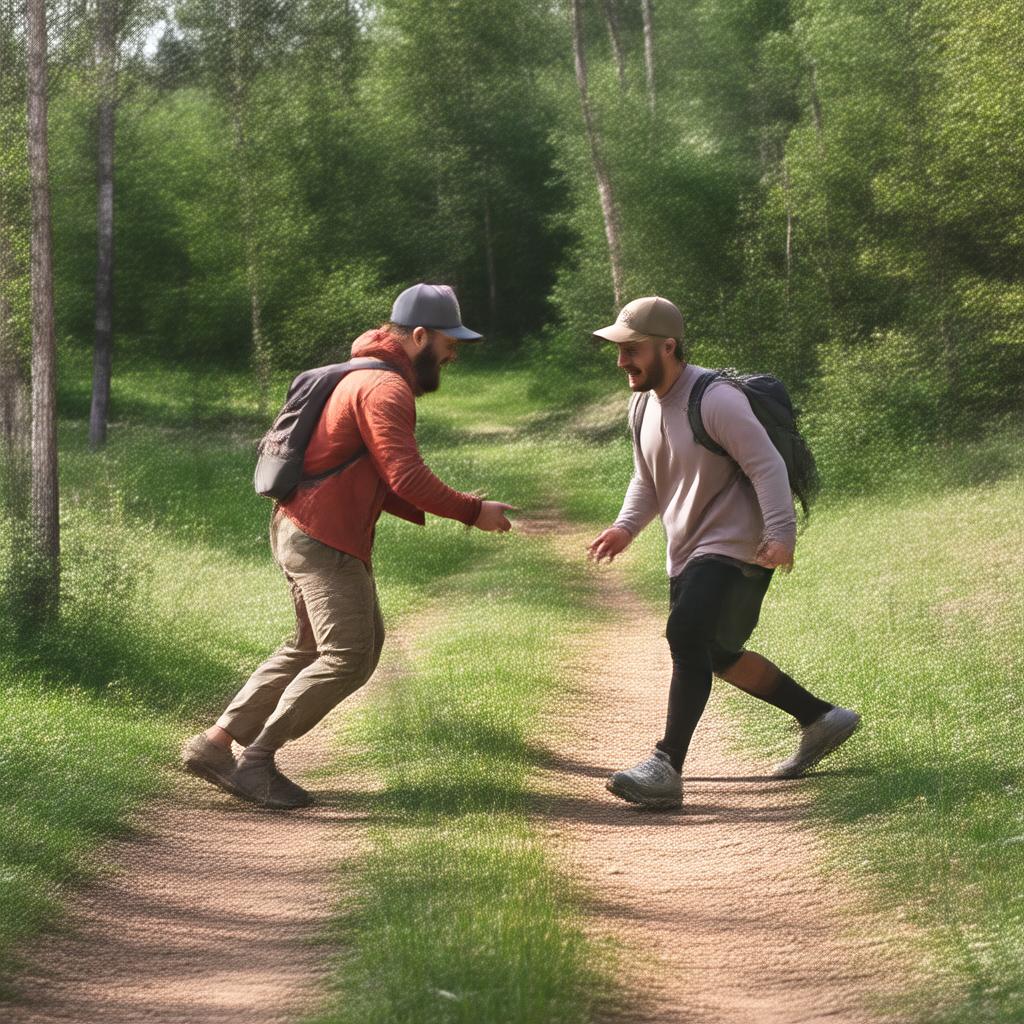}
        \caption*{A-Euler}
    \end{subfigure}
    \caption{Prompt: Two men play around on a nature trail}
    \label{fig:paired_comparisons}
\end{figure}

\end{document}